\documentclass[sigconf]{acmart}

\setcopyright{none}
\renewcommand\footnotetextcopyrightpermission[1]{}
\acmConference[Under Review]{Under Review}{2027}{Anonymous}
\acmYear{2027}

\usepackage{enumitem}

\usepackage{tikz}
\usepackage{xcolor}
\usepackage{amsmath,amssymb}
\usepackage{graphicx}
\usetikzlibrary{positioning,calc,fit,arrows.meta}

\usepackage{amsmath}
\usepackage{amssymb}
\usepackage{booktabs}
\usepackage{multirow}
\usepackage{graphicx}
\usepackage{colortbl}
\usepackage{soul}
\usepackage{microtype}
\usepackage{placeins}
\usepackage{cuted}
\usepackage{xspace}
\usepackage{booktabs}
\newtheorem{proposition}{Proposition}
\newcommand{\method}{\textsc{ADRS}\xspace}
\newcommand{\tva}{\textsc{TVA}\xspace}
\newcommand{\sg}{\operatorname{sg}}
\definecolor{topcolor}{HTML}{E2F0D9}
\definecolor{secondcolor}{HTML}{FFF2CC}
\definecolor{thirdcolor}{HTML}{DDEBF7}

\begin{document}
\title{Agentic Reinforcement Learning with Self-Distilled Reward Shaping}

\author{Ranxu Zhang}
\authornote{This work was conducted during an internship at Alibaba Group.}
\affiliation{%
  \institution{University of Science and Technology of China}
  \city{Hefei}
  \country{China}
}

\author{Guinan Chen}
\affiliation{%
  \institution{University of Science and Technology of China}
  \city{Hefei}
  \country{China}
}

\author{Chenshaodong}
\affiliation{%
  \institution{Alibaba Group}
  \city{Hangzhou}
  \country{China}
}

\author{Jinghao Lin}
\affiliation{%
  \institution{Alibaba Group}
  \city{Hangzhou}
  \country{China}
}

\author{Xiaozhou Xu}
\affiliation{%
  \institution{Alibaba Group}
  \city{Hangzhou}
  \country{China}
}

\author{sunzhe}
\affiliation{%
  \institution{Alibaba Group}
  \city{Hangzhou}
  \country{China}
}

\author{Yanyong Zhang}
\affiliation{%
  \institution{University of Science and Technology of China}
  \city{Hefei}
  \country{China}
}

\author{Chao Wang}
\authornote{Chao Wang is the corresponding author.}
\affiliation{%
  \institution{University of Science and Technology of China}
  \city{Hefei}
  \country{China}
}

\renewcommand{\shortauthors}{Zhang et al.}

\begin{abstract}
Agentic reinforcement learning enables LLM agents to learn through interaction, but sparse trajectory-level rewards reveal success without identifying which intermediate decisions deserve credit. Training-only privileged skills can provide denser supervision by allowing the same frozen policy snapshot to rescore fixed tokens from skill-free trajectories while conditioned on task-matched procedural skills. Existing methods, however, do not jointly calibrate teacher scores across interaction steps, relate teacher confidence to realized returns, and integrate the resulting signal into native reward-to-advantage construction. We introduce \textbf{A}gentic Reinforcement Learning with Self-\textbf{D}istilled \textbf{R}eward \textbf{S}haping (\method), a framework for constructing return-associated token-level credit for multi-turn language agents. \method centers and normalizes privileged token scores within each step, modulates them with a return-associated Teacher Value Advantage (\tva) gate based on within-group confidence--return association, and incorporates the gated token signal into native RL credit construction. Together, these components determine what the teacher prefers, when that preference is return-relevant, and how it enters the native reinforcement-learning credit path, while keeping rollouts and inference skill-free. Finally, experiments across three interactive benchmarks show that \method consistently improves performance on long-horizon tasks, with gains persisting across RL backbones, reduced-data settings, unseen tasks, and extended training. Our code is available at the following the link: https://github.com/gitrxh/ADRS-arxiv
\end{abstract}



\maketitle
\input{figures/challenge1}
\section{Introduction}

Language-model agents can solve interactive tasks involving evidence search~\cite{jin2025search}, embodied navigation~\cite{shridhar2020alfworld}, and web interaction~\cite{yao2022webshop}. However, agentic reinforcement learning typically provides only terminal outcome rewards, which indicate whether a trajectory succeeds but not which intermediate query or action deserves credit. This sparse feedback makes token-level temporal credit assignment difficult even for critic-free group-relative methods such as GRPO~\cite{shao2024deepseekmath} and GiGPO~\cite{feng2026group}. Natural-language skills can provide denser guidance by describing subgoals, action rules, and common failure patterns~\cite{wang2026skill,lu2026self,xia2026skillrl}. We therefore consider a training-only privileged setting in which the same frozen policy snapshot rescores fixed tokens from skill-free rollouts under task-matched skills; skills remain absent at inference.

This setting connects knowledge distillation~\cite{hinton2015distilling} with learning using privileged information~\cite{vapnik2009new}. Building on on-policy distillation~\cite{agarwal2024policy}, OPSD~\cite{zhao2026self} established a basic recipe in which the same model uses privileged context to supervise trajectories generated without that context. Recent OPSD+RL methods incorporate this supervision in different ways: GRPO+OPSD directly combines the distillation and policy-optimization objectives; Skill-SD~\cite{wang2026skill} extracts dynamic skills from completed trajectories; RLSD~\cite{yang2026self} adjusts token-level updates using teacher--student differences; and SDAR~\cite{lu2026self} introduces retrieved procedural skills with confidence-gated auxiliary distillation. Although these methods enrich the source and weighting of privileged supervision, a central question remains: how can privileged token confidence be associated with realized returns and converted into temporal credit through a single native reward-to-advantage-to-policy path?

This question decomposes into three technical challenges as shown in Figure~\ref{fig:effective_reward_utilization}: score calibration, reliability estimation, and credit integration. First, raw teacher log-probabilities depend on local context, policy uncertainty, and response composition. Their offsets and scales are therefore not directly comparable across interaction steps. Second, a confident teacher is not necessarily a useful teacher: high token likelihood may reflect linguistic familiarity rather than decisions associated with higher environment returns. Third, a separate distillation objective places privileged supervision outside the backbone's reward-to-advantage computation and may misalign token-level updates with the group-relative advantages constructed by GRPO or GiGPO. Addressing these challenges requires a mechanism that calibrates teacher scores across steps, estimates their association with realized returns, and injects the resulting token signal before advantage construction.

To this end, we introduce \method, a self-distilled reward-shaping framework that transforms task-matched procedural skills into token-level training rewards while keeping agent rollouts and inference skill-free. \method addresses the three challenges through three coupled designs. First, it re-scores student-generated tokens under the same behavior-policy snapshot with privileged skill context, then applies within-step centering and scale normalization to retain relative token preferences while suppressing step-dependent drift. Second, a selected-level return-associated Teacher Value Advantage (\tva) gate modulates the normalized signal using the within-group association between privileged confidence and realized return, rather than assuming uniform teacher reliability. Third, the gated signal is added to the base token rewards before the configured advantage operator, keeping privileged guidance within a single reward-to-advantage-to-policy path rather than a separate auxiliary objective. Together, these components determine what the privileged teacher prefers, when that preference is return-relevant, and how it enters the native reinforcement-learning credit path. Our contributions are:
\begin{itemize}[leftmargin=1.2em,labelsep=0.35em,topsep=0.5pt,partopsep=0pt,itemsep=0pt,parsep=0pt]

\item We propose \method, which converts task-matched procedural skills into return-associated token rewards through within-step score calibration and \tva gating, while keeping rollouts and inference skill-free. For the centered L2/L3 variants, we further establish a stepwise zero-sum structure.
\item We formulate the gated signal as pre-advantage reward shaping, integrating privileged guidance into the native reward-to-advantage-to-policy path rather than a separate auxiliary objective. We characterize its compatibility with GRPO/GiGPO-style credit operators and its local first-order connection to sampled-token auxiliary updates.
\item Experiments across three interactive benchmarks show that \method delivers consistent performance gains on long-horizon tasks across RL backbones, remains effective with reduced training data, generalizes to unseen tasks, and sustains its gains over extended training.
\end{itemize}

\vspace{-1ex}
\section{Related Work}

\subsection{Reinforcement Learning for LLMs}
Reinforcement learning provides the optimization backbone for training LLMs from environment feedback. PPO~\cite{schulman2017proximal} uses a clipped policy objective, whereas GRPO~\cite{shao2024deepseekmath} removes the critic through within-group outcome normalization. For multi-turn interaction, LOOP~\cite{chen2025reinforcement} develops value-free training for interactive tasks, while RAGEN~\cite{wang2025ragen} examines instability and degeneracy during optimization. As interaction horizons grow, localizing outcome feedback becomes increasingly important. GiGPO~\cite{feng2026group} introduces episode- and step-level relative advantages, while Agent Lightning~\cite{luo2025agent} decomposes trajectories into trainable transitions. 
These methods improve temporal credit assignment but remain driven by sparse outcomes. ADRS supplements them with privileged
token-level guidance before advantage computation.

\vspace{-1ex}
\subsection{Privileged Self-Distillation}

Privileged self-distillation combines teacher--student transfer with information available only during training. Knowledge distillation transfers teacher knowledge to a student~\cite{hinton2015distilling}, while privileged-information learning provides additional training-only supervision~\cite{vapnik2009new}. On-policy distillation, DAgger-style learner-distribution supervision, and context distillation extend these ideas to student-generated states and richer contexts~\cite{agarwal2024policy,ross2011reduction,snell2022learning}. OPSD~\cite{zhao2026self} brings them together by constructing teacher and student views from the same model under different contexts.


Recent methods combine distillation with reward-based learning
through SDPO \cite{hubotter2026reinforcement} and KDRL
\cite{xu2025kdrl}, or stabilize and selectively apply supervision
through TCOD \cite{wang2026tcod}, HDPO \cite{ding2026hdpo},
and TIP \cite{xu2026tip}. OPCD \cite{ye2026policy} instead
focuses on internalizing enriched context along student-generated
trajectories. In language-agent settings, Skill-SD
\cite{wang2026skill} uses trajectory-derived skills, RLSD
\cite{yang2026self} uses teacher--student differences to modulate
token updates, and SDAR \cite{lu2026self} uses gated retrieved
skills. The differences between SDAR and ADRS are summarized in
Table~\ref{tab:related}. Self-Distilled Policy Gradient \cite{liu2026self},
Rebellious Student \cite{kim2026rebellious}, and CRAFT
\cite{meng2026craft} further investigate full-vocabulary
supervision, reversed teacher signals, and counterfactual credit
from sibling rollouts, respectively. In contrast, ADRS converts
calibrated, return-associated teacher scores into token rewards
before native advantage construction.

\begin{table*}[t]
\caption{Positioning relative to the closest privileged self-teaching
methods. ``Backward'' refers to the policy update.}
\vspace{-10pt}
\label{tab:related}
\centering
\small
\setlength{\tabcolsep}{4.5pt}
\begin{tabular}{lllll}
\toprule
Method & Skill source & Teacher signal & Injection point & Policy backward \\
\midrule
SDAR~\cite{lu2026self}
& Retrieved external SkillBank
& Gated teacher-student token gap
& Auxiliary distillation loss
& RL + auxiliary term \\

\method (ours)
& Task-indexed privileged provider
& Centered teacher token score + \tva
& Native RL credit construction
& Single RL objective \\
\bottomrule
\end{tabular}
\vspace{-2ex}
\end{table*}
\vspace{-2ex}
\subsection{Reward Shaping and Credit Assignment}

Reward shaping supplements sparse feedback with denser signals. Potential-based shaping can preserve optimal policies when based on a fixed Markov potential~\cite{ng1999policy}, while RUDDER~\cite{arjona2019rudder} redistributes delayed returns toward outcome-relevant decisions. Process reward models supervise intermediate reasoning steps~\cite{lightman2024let}, but require a separately trained verifier and process-level supervision. Unlike separate reward models or auxiliary distillation objectives, ADRS derives token-level rewards from a privileged policy branch and injects them before advantage construction.

\section{Problem Formulation}
\label{sec:problem}

In multi-turn language-agent reinforcement learning, each task $x_i\sim\mathcal D$ induces a partially observed interaction with an environment. Without privileged text, a behavior-policy snapshot $\pi_{\theta_b}$ samples $K$ trajectories $\{\tau_{i,k}\}_{k=1}^{K}$ for each task. At step $s$, the history $h_{i,k,s}$ contains the task, observation, and prior actions, and the policy emits $y_{i,k,s}=(y_{i,k,s,1},\ldots,y_{i,k,s,L_{i,k,s}})$ with valid token set $\mathcal V_{i,k,s}$. Environment feedback is mapped to base token rewards $r^{\mathrm{base}}$ under the same valid-token mask used by teacher scoring, advantage estimation, and the actor loss. For $\mathcal B\in\{\mathrm{GRPO},\mathrm{GiGPO}\}$~\cite{shao2024deepseekmath,feng2026group}, let $A^{\mathcal B}=\operatorname{Adv}_{\mathcal B}(r^{\mathrm{base}})$ denote the native token-aligned advantage, including the backbone's aggregation, discounting, grouping, normalization, and broadcasting rules.


In each training iteration, $\pi_{\theta_b}$ first performs the
ordinary rollout and receives environment feedback. After the
trajectories are fixed, a deterministic provider returns
task-matched procedural knowledge $\rho(x_i)$, and the same frozen
snapshot rescores the realized tokens under
$\rho(x_i)\oplus h_{i,k,s}$. This branch changes only the context:
it neither samples another trajectory nor introduces a separate
or larger teacher, and privileged text is absent from student
rollouts and inference.
The paired scores, environment returns, and grouping metadata
provide training-only evidence for refining credit among the
realized token decisions. Our goal is to use this evidence to
assign token-level credit to return-relevant intermediate
decisions, thereby improving the skill-free policy $\pi_\theta$
while preserving the native GRPO/GiGPO trajectory direction and requiring no privileged input during rollout or inference.
\section{Method}
\label{sec:method}

\begin{figure*}[t]
    \centering
    \includegraphics[width=\textwidth]{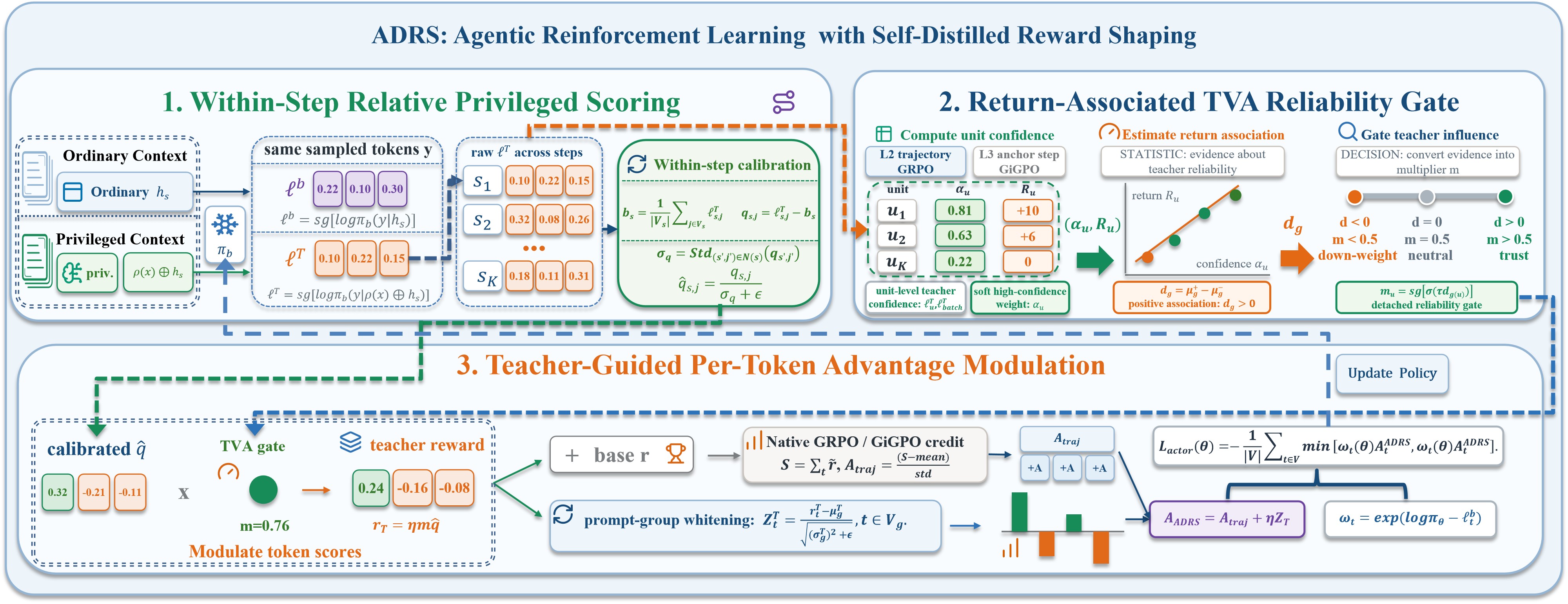}
    \vspace{-20pt}
    \Description{Flow diagram of ADRS. A student rollout produces base token rewards. A detached behavior-policy snapshot reads task-matched skill text and scores student tokens. Raw teacher confidence and observed returns feed TVA, while centered teacher scores form the dense signal. The gated score is added to the base reward before group-relative reinforcement-learning backbones such as GRPO and GiGPO estimate advantages, after which the clipped policy update is applied.}
    \caption{\method training flow. A behavior policy samples without privileged text. The same frozen snapshot re-scores those tokens under a task-matched skill. Step-centered scores provide the dense signal, while observed returns calibrate its optional \tva gate. The gated signal is added to the base reward before group-relative backbones such as GRPO and GiGPO estimate advantages; a standard clipped actor update then follows.}
    \label{fig:overview}
\end{figure*}

ADRS addresses how to convert training-only privileged guidance into return-associated token credit without changing skill-free rollout and inference or replacing the underlying RL backbone. As shown in Figure~\ref{fig:overview}, it operates as a credit-construction layer between trajectory collection and actor optimization. In each training iteration, the behavior policy first samples $K$ trajectories without skill text and receives environment outcomes and base token rewards. Once these trajectories are fixed, the same frozen policy snapshot re-scores the realized tokens under ordinary and privileged contexts. The paired scores and observed outcomes provide the evidence needed to determine which tokens receive stronger privileged support, whether that support is associated with return, and how it should affect the subsequent policy update.

These three questions map directly to three modules. Module~1, \emph{Within-Step Relative Privileged Scoring}, calibrates privileged token scores within each interaction step to obtain comparable relative preferences. Module~2, \emph{Return-Associated TVA Reliability Gating}, uses the raw privileged scores and observed returns to estimate teacher reliability within the selected comparison group. Because Module~2 does not depend on the calibrated output of Module~1, the two modules operate in parallel and meet only in Module~3. Module~3, \emph{Teacher-Guided Per-Token Advantage Modulation}, uses the reliability estimate to modulate the calibrated token signal and combines the resulting token-level redistribution with the environment-defined trajectory direction. The resulting detached advantage is passed to the standard GRPO/GiGPO actor update, after which the updated actor becomes the behavior policy for the next iteration. Privileged text is used only for post-rollout rescoring and remains absent during rollout and inference.

\subsection{Within-Step Relative Privileged Scoring}

ADRS first considers the comparability of privileged scores across interaction steps. Teacher log-probabilities undergo global shifts and scale changes with context length, interaction position, and response composition. A token with a high raw score in one step can therefore carry less relative support than a token with a lower raw score in another. To obtain a stable token signal, ADRS treats privileged scores as relative evidence inside each realized decision step: it removes the within-step offset, normalizes dispersion, and outputs the signed, scale-controlled score $\widehat q_{s,j}$. Raw privileged scores are retained for the subsequent return-association estimate.

For each sampled token, the ordinary-context and privileged-context behavior-policy scores are
\begin{equation}
\small
\begin{aligned}
\ell^b_{s,j}
&=\sg\!\left[\log \pi_{\theta_b}\!\left(y_{s,j}\mid h_s,y_{s,<j}\right)\right],
\\
\ell^T_{s,j}
&=\sg\!\left[\log \pi_{\theta_b}\!\left(y_{s,j}\mid \rho(x)\oplus h_s,y_{s,<j}\right)\right].
\end{aligned}
\label{eq:teacher-logprob}
\end{equation}
The snapshot $\theta_b$ generates the batch and remains frozen during both rescoring passes. Thus, $\ell^T$ measures the support that the privileged context assigns to a token sampled by the student policy; all scores are defined on the same realized student trajectory.

The centered and normalized score is
\begin{equation}
\small
\begin{aligned}
b_s
&=\frac{1}{|\mathcal V_s|}\sum_{j\in\mathcal V_s}\ell^T_{s,j},
&
q_{s,j}
&=\ell^T_{s,j}-b_s,
\\
\sigma_q
&=\operatorname{Std}_{(s',j')\in\mathcal N(s)}\!\left(q_{s',j'}\right),
&
\widehat q_{s,j}
&=\frac{q_{s,j}}{\sigma_q+\epsilon}.
\end{aligned}
\label{eq:teacher-score}
\end{equation}
Here $\epsilon>0$ and $\mathcal N(s)$ is the chosen normalization scope. The denominator is shared within step $s$, giving
\begin{equation}
\small
\sum_{j\in\mathcal V_s}\widehat q_{s,j}=0.
\label{eq:zero-sum}
\end{equation}
Within-step centering removes the overall confidence offset specific to an interaction position, while $\sigma_q$ makes the strength of teacher guidance robust to changes in score dispersion. A positive $\widehat q_{s,j}$ indicates that a sampled token receives a privileged score above the step average. ADRS therefore identifies which tokens within the same decision step should receive relatively stronger or weaker credit; the next subsection evaluates the consistency of this relative support with environment return.

\subsection{Return-Associated TVA Reliability Gate}

ADRS next considers the reliability of privileged confidence. High teacher confidence becomes useful RL guidance when it aligns with observed return in comparable units. To evaluate this alignment, ADRS compares soft teacher confidence with pre-ADRS return inside a selected group and produces a scalar gate for its tokens. This reliability operation controls teacher influence before the final credit-construction operation.

TVA supports two granularities. Completion-level TVA (L2) matches the comparison available in GRPO: trajectories sampled for the same prompt form the group, and each trajectory uses its pre-ADRS base return. Step-level TVA (L3) matches GiGPO's additional structure: repeated anchor states define comparable interaction steps, and each step uses return-to-go from that state. In this sense, L2 uses prompt-level trajectory information, while L3 uses the anchor-state and step-return information introduced by GiGPO. A training run selects one level.

Let $\mathcal U$ denote the units at the selected level and $\mathcal V_u$ the valid tokens of unit $u$. The reliability operation first computes unit confidence and a soft high-confidence weight:
\begin{equation}
\footnotesize
\bar\ell^T_u=\frac{1}{|\mathcal V_u|}\sum_{(s,j)\in\mathcal V_u}\ell^T_{s,j},
\qquad
\bar\ell^T_{\mathrm{batch}}=\frac{1}{|\mathcal U|}\sum_{u\in\mathcal U}\bar\ell^T_u,
\qquad
\alpha_u=\sigma\!\left(\frac{\bar\ell^T_u-\bar\ell^T_{\mathrm{batch}}}{T}\right).
\label{eq:tva-confidence}
\end{equation}
It then contrasts returns on the high- and low-confidence sides:
\begin{equation}
\footnotesize
\mu_g^+=\frac{\sum_{u\in g}\alpha_uR_u}{\sum_{u\in g}\alpha_u+\epsilon},
\qquad
\mu_g^-=\frac{\sum_{u\in g}(1-\alpha_u)R_u}{\sum_{u\in g}(1-\alpha_u)+\epsilon},
\qquad
d_g=\mu_g^+-\mu_g^-.
\label{eq:tva-contrast}
\end{equation}
After optional positive scale normalization to $\widetilde d_g$, the detached gate is
$
\small
m_u=\sg\!\left[\sigma\!\left(\tau\widetilde d_{g(u)}\right)\right].
\label{eq:tva}
$
The gate is broadcast to all tokens in unit $u$. Positive return association increases it above $0.5$, negative association lowers it, undersized groups use $0.5$, and the TVA-off configuration uses $m=1$.

\begin{proposition}[TVA as return association]
\label{prop:tva-covariance}
Ignoring the numerical $\epsilon$ in Eq.~\ref{eq:tva-contrast}, let $\bar\alpha_g=\mathbb E_g[\alpha]$. For any non-degenerate group,
\begin{equation}
\small
d_g=\frac{\operatorname{Cov}_g(\alpha,R)}{\bar\alpha_g(1-\bar\alpha_g)}.
\label{eq:tva-covariance}
\end{equation}
Thus, the sign of TVA agrees with the within-group empirical covariance between privileged confidence and return.
\end{proposition}

Proposition~\ref{prop:tva-covariance} gives the gate its statistical meaning. Since the denominator is positive, $d_g$ increases exactly in the intended direction: units with higher privileged confidence also tend to have higher returns inside the comparison group. This explains why ADRS can use a group-level gate while allowing pairwise teacher errors to be absorbed by the group statistic. The proposition is observational and batch-local; finite $\epsilon$ gives the stabilized implementation. The resulting gate $m$ is then combined with the calibrated token score in the final credit-construction operation.

\subsection{Teacher-Guided Per-Token Advantage Modulation}
\label{sec:token-advantage}

ADRS finally considers how the calibrated and reliability-weighted teacher signal should enter policy credit. The within-step score is zero-sum, so a standard trajectory sum preserves outcome ordering while removing the centered signal. ADRS therefore adopts two complementary paths. The trajectory path carries the environment-defined direction of each sampled completion. The token path saves and whitens the privileged signal so that update strength can vary across tokens within the same trajectory.

The teacher reward and shaped token reward are
\begin{equation}
\small
r^T_{s,j}=\eta m_{s,j}\widehat q_{s,j},
\qquad
\widetilde r_{s,j}=r^{\mathrm{base}}_{s,j}+r^T_{s,j},
\label{eq:star-reward}
\end{equation}
where $\eta\ge0$ controls the teacher scale. We keep $\eta$ in $r^T$ so that the saved tensor matches the injected reward. For positive global $\eta$, whitening removes this global scale from $Z^T$, making the outer $\eta$ in Eq.~\ref{eq:star-advantage} the effective token-modulation coefficient.
\vspace{-3pt}
\begin{proposition}[Stepwise zero-sum representation]
\label{prop:centered-potential}
If the normalization denominator and gate are shared within interaction step $s$, the teacher reward in Eq.~\ref{eq:star-reward} satisfies
$\sum_{j\in\mathcal V_s}r^T_{s,j}=0$.
For any ordering $(j_1,\ldots,j_{n_s})$ of the valid tokens, it admits a sampled-chain representation
$r^T_{s,j_r}=\Phi_{s,r+1}-\Phi_{s,r}$ with
$\Phi_{s,1}=\Phi_{s,n_s+1}=0$.
This identity is an algebraic representation on the sampled step; policy-invariance would require a fixed Markov potential.
\end{proposition}

The proposition explains this two-path credit construction. The teacher reward adds zero step-level mass under the stated sharing condition, so trajectory summation keeps the environment ordering. The same zero-sum identity motivates a token-preserving path for training the privileged structure. ADRS therefore sends $\widetilde r$ through the trajectory path and saves $r^T$ for token modulation.

For GRPO, the trajectory path computes
\begin{equation}
\small
\begin{aligned}
S_{i,k}
&=\sum_{t\in\mathcal V_{i,k}}\widetilde r_{i,k,t},
&
\mu_i^S
&=\operatorname{Mean}_{k=1}^{K}(S_{i,k}),
\\
(\sigma_i^S)^2
&=\operatorname{Var}_{k=1}^{K}(S_{i,k}),
&
A^{\mathrm{traj}}_{i,k}
&=\frac{S_{i,k}-\mu_i^S}{\sqrt{(\sigma_i^S)^2+\epsilon}}.
\end{aligned}
\label{eq:trajectory-advantage}
\end{equation}
Because the step-shared teacher reward sums to zero, $S_{i,k}$ equals the base trajectory score in this configuration; $A^{\mathrm{traj}}$ carries the trajectory-level outcome direction.

The token path whitens the saved teacher reward within prompt group $g$:
\vspace{-10pt}
\begin{equation}
\small
\begin{aligned}
\mu_g^T
&=\operatorname{Mean}_{v\in\mathcal V_g}(r^T_v),
&
(\sigma_g^T)^2
&=\operatorname{Var}_{v\in\mathcal V_g}(r^T_v),
\\
Z^T_t
&=\frac{r^T_t-\mu_g^T}{\sqrt{(\sigma_g^T)^2+\epsilon}},
&
t&\in\mathcal V_g.
\end{aligned}
\label{eq:teacher-modulation}
\end{equation}
The final token-level advantage is
\begin{equation}
\small
A^{\mathrm{ADRS}}_{i,k,t}
=A^{\mathrm{traj}}_{i,k}+\eta Z^T_{i,k,t}.\label{eq:star-advantage}
\end{equation}

The first term sets trajectory direction; the second redistributes update strength along it. Since $Z^T$ is standardized, $\eta$ directly controls token modulation and recovers the native algorithm at $\eta=0$. For GiGPO, ADRS adds $Z^T$ to the episode-level advantage while retaining the native step-level branch.

For flattened valid token $t$ with ordinary context $c_t$, define
\begin{equation}
\footnotesize
\omega_t(\theta)
=\exp\!\left(\log\pi_\theta(y_t\mid c_t)-\ell^b_t\right),
\qquad
\bar\omega_t(\theta)
=\operatorname{clip}\!\left(\omega_t(\theta),1-\varepsilon_{\mathrm{lo}},1+\varepsilon_{\mathrm{hi}}\right).
\label{eq:policy-ratio}
\end{equation}
With detached $A^{\mathrm{ADRS}}$, the actor uses
\begin{equation}
\small
\mathcal L_{\mathrm{actor}}(\theta)
=-\frac{1}{|\mathcal V|}\sum_{t\in\mathcal V}
\min\!\left[
\omega_t(\theta)A^{\mathrm{ADRS}}_t,
\bar\omega_t(\theta)A^{\mathrm{ADRS}}_t
\right].
\label{eq:star-objective}
\end{equation}
Thus, privileged information affects learning through the ordinary policy ratio after being converted into detached credit.

\begin{proposition}[Trajectory preservation and local token-gradient equivalence]
\label{prop:unified-gradient}
Under a step-shared gate, Eq.~\ref{eq:star-reward} leaves the GRPO trajectory score, and hence $A^{\mathrm{traj}}$, unchanged. If the teacher signal is non-constant within the group, Eq.~\ref{eq:star-advantage} nevertheless induces nonzero token-level credit. Define $A_t^0=A^{\mathrm{traj}}_{i,k}$ and $\Delta A_t=\eta Z^T_t$. At the behavior-policy point with clipping inactive,
\begin{equation}
\footnotesize
\left.\nabla_\theta J_{\mathrm{ADRS}}\right|_{\theta_b}
=\frac{1}{|\mathcal V|}\sum_{t\in\mathcal V}
(A_t^0+\Delta A_t)\nabla_\theta\log\pi_\theta(y_t\mid c_t).
\label{eq:star-score-update}
\end{equation}
For a detached sampled-token auxiliary coefficient $a_t$, the corresponding local update is
\begin{equation}
\small
\left.\nabla_\theta J_{\mathrm{aux}}\right|_{\theta_b}
=\frac{1}{|\mathcal V|}\sum_{t\in\mathcal V}
(A_t^0+a_t)\nabla_\theta\log\pi_\theta(y_t\mid c_t).
\label{eq:aux-score-update}
\end{equation}
Setting $a_t=\Delta A_t=\eta Z^T_t$ makes the two first-order policy gradients identical at this evaluation point.
\end{proposition}
\vspace{-3pt}
Proposition~\ref{prop:unified-gradient} clarifies the optimization role of ADRS. The trajectory path carries the return-defined decision, and the token path contributes a local score-weighted update equivalent to a detached token auxiliary coefficient at the behavior-policy point. The statement is local to the sampled batch and unclipped branch; broader objective equivalences require separate assumptions. ADRS then passes $A^{\mathrm{ADRS}}$ to GRPO/GiGPO, closing the iteration and preparing the next behavior snapshot.

\section{Experiments}
\label{sec:experiments}
\subsection{Experimental Settings}
\label{sec:setup}

\paragraph{Benchmarks.}
We evaluate \method across three complementary language-agent interactions. ALFWorld~\cite{shridhar2020alfworld} casts household tasks as text-based embodied control, requiring agents to satisfy ordered preconditions over extended action sequences. WebShop~\cite{yao2022webshop} evaluates web navigation, where an agent searches, inspects product attributes, retains user constraints, and completes a purchase. Search-based QA follows Search-R1~\cite{jin2025search} and tests evidence gathering over seven question-answering datasets using only four tool-use turns. The combination separates long-horizon procedural control from shorter information-seeking interaction, allowing us to test where privileged token credit is most useful. Appendix~\ref{app:benchmarks} provides dataset-level descriptions, split usage, metrics, and interaction limits.
\paragraph{Baselines.}
We organize comparisons by supervision source and use. Vanilla and Skill-Prompt test the instruction-tuned policy without or with inference-time skills. GRPO~\cite{shao2024deepseekmath}, Skill-GRPO, and GiGPO~\cite{feng2026group} represent outcome-driven reinforcement learning, with GiGPO supplying a step-aware credit backbone. OPSD~\cite{zhao2026self}, GRPO+OPSD, Skill-SD~\cite{wang2026skill}, RLSD~\cite{yang2026self}, and SDAR~\cite{lu2026self} provide structured self-distillation or skill-conditioned supervision. These groups distinguish gains from outcome optimization, skill access, auxiliary token supervision, and native RL credit construction. Unless marked by an asterisk, evaluation uses no skill context; \method never uses privileged skills during rollout or inference. Appendix~\ref{app:baselines} provides details.
\paragraph{Models, metrics, and protocol.}
The common 150-step comparison covers Qwen2.5-3B-Instruct, Qwen2.5-7B-Instruct~\cite{qwen2025qwen25technicalreport}, and Qwen3-1.7B-Instruct~\cite{yang2025qwen3}. We report ALFWorld task-family and overall success, Search per-dataset exact match and macro-average, and WebShop normalized score and exact success. We extend representative runs to 300 steps, separate GRPO and GiGPO backbones, inspect action/object token gaps, vary available training data, evaluate unseen transfer, and sweep maximum interaction horizon against SDAR. All experiments follow the SDAR benchmark and evaluation setup~\cite{lu2026self}, using multiple NVIDIA H100 GPUs. Appendix~\ref{app:implementation} provides additional implementation details.

\subsection{Overall Performance}
\label{sec:overall-results}

\begin{table*}[ht!]
    \centering
    \caption{
        \textbf{Performance Comparison on the representative long-horizon benchmarks (ALFWorld, Search-based QA, and WebShop).}
        We report the success rate (\%) on ALFWorld, accuracy on search-based QA, and task-completion score/success rate on WebShop. An asterisk (*) denotes validation with skills. The \sethlcolor{topcolor}\hl{\textbf{best}}, \sethlcolor{secondcolor}\hl{\mbox{\underline{second-best}}}, and \sethlcolor{thirdcolor}\hl{third-best} results are highlighted.
    }
    \vspace{-10pt}
    \label{tab:main_results_no_opid}
    \resizebox{1\textwidth}{!}{%
    \begin{tabular}{l ccccccc cccccccc cc}
    \toprule
    & \multicolumn{7}{c}{\textbf{ALFWorld}} & \multicolumn{8}{c}{\textbf{Search-based QA}} & \multicolumn{2}{c}{\textbf{WebShop}} \\
    \cmidrule(lr){2-8} \cmidrule(lr){9-16} \cmidrule(lr){17-18}
    \textbf{Method}
    & \textbf{Pick} & \textbf{Look} & \textbf{Clean} & \textbf{Heat} & \textbf{Cool} & \textbf{Pick2} & \textbf{Avg}
    & \textbf{NQ} & \textbf{Triv} & \textbf{Pop} & \textbf{Hotp} & \textbf{2Wk} & \textbf{MuS} & \textbf{Bam} & \textbf{Avg}
    & \textbf{Score} & \textbf{Succ.} \\
    \midrule
    \rowcolor{gray!10} \multicolumn{18}{l}{\textit{Qwen2.5-3B-Instruct}} \\
    Vanilla
        & 44.4 & 11.1 & 6.2 & 15.4 & 28.6 & 12.5 & 21.9
        & 24.6 & 48.1 & 31.0 & 26.3 & 25.3 & 7.2 & 59.7 & 31.7
        & 6.7 & 0.8
        \\
    Skill-Prompt*
        & 51.7 & 66.7 & 48.4 & 0.0 & 4.3 & 10.0 & 28.9
        & 23.7 & 46.2 & 30.6 & 24.4 & 22.1 & 7.5 & 12.5 & 23.9
        & 0.2 & 0.8
        \\
    OPSD
        & 48.8 & 41.7 & 16.7 & 0.0 & 15.8 & 16.7 & 28.1
        & 0.1 & 0.1 & 0.1 & 0.0 & 0.0 & 0.0 & 0.0 & 0.0
        & 11.3 & 3.1
        \\
    GRPO
        & 91.2 & 62.5 & 96.2 & 61.9 & 65.0 & 47.4 & 75.0
        & 39.3 & \cellcolor{thirdcolor}60.6 & 41.1 & 37.4 & 34.6 & 15.4 & 26.4 & 36.4
        & 79.8 & 63.3
        \\
    Skill-GRPO
        & 88.9 & 71.4 & 58.8 & 70.6 & 40.7 & 29.2 & 60.2
        & 43.5 & 58.8 & 43.0 & 36.8 & 32.2 & 11.7 & 12.5 & 34.1
        & 77.3 & 60.9
        \\
    Skill-GRPO*
        & \cellcolor{thirdcolor}94.3 & 57.1 & \cellcolor{topcolor}\textbf{100.0} & 66.7 & \cellcolor{thirdcolor}73.1 & 57.1 & 80.5
        & 44.3 & 59.6 & \cellcolor{thirdcolor}44.3 & 39.0 & 36.1 & 14.5 & 14.9 & 36.1
        & 76.3 & \cellcolor{thirdcolor}66.4
        \\
    GRPO+OPSD
        & \cellcolor{topcolor}\textbf{100.0} & \cellcolor{topcolor}\textbf{82.4} & 85.7 & \cellcolor{secondcolor}\underline{75.0} & 70.0 & 60.0 & \cellcolor{thirdcolor}81.2
        & \cellcolor{topcolor}\textbf{44.9} & \cellcolor{secondcolor}\underline{61.2} & \cellcolor{secondcolor}\underline{45.2} & \cellcolor{topcolor}\textbf{40.4} & \cellcolor{thirdcolor}38.5 & \cellcolor{secondcolor}\underline{16.0} & \cellcolor{secondcolor}\underline{66.1} & \cellcolor{secondcolor}\underline{44.6}
        & 77.8 & \cellcolor{thirdcolor}66.4
        \\
    Skill-SD
        & 88.2 & 50.0 & 96.2 & 52.4 & 65.0 & 57.9 & 73.4
        & 44.4 & 60.4 & 44.0 & 39.5 & \cellcolor{topcolor}\textbf{40.4} & 15.4 & 64.9 & \cellcolor{thirdcolor}44.1
        & 75.9 & 64.0
        \\
    RLSD
        & 87.9 & \cellcolor{secondcolor}\underline{75.0} & 90.9 & \cellcolor{secondcolor}\underline{75.0} & \cellcolor{thirdcolor}73.1 & \cellcolor{thirdcolor}68.4 & 79.7
        & 41.5 & 58.6 & 42.3 & \cellcolor{topcolor}\textbf{40.4} & \cellcolor{secondcolor}\underline{40.2} & \cellcolor{topcolor}\textbf{16.8} & \cellcolor{topcolor}\textbf{66.9} & 43.8
        & \cellcolor{thirdcolor}84.4 & \cellcolor{thirdcolor}66.4
        \\
    SDAR
        & \cellcolor{secondcolor}\underline{97.1} & 62.5 & \cellcolor{topcolor}\textbf{100.0} & 61.9 & \cellcolor{secondcolor}\underline{75.0} & \cellcolor{secondcolor}\underline{84.2} & \cellcolor{secondcolor}\underline{84.4}
        & \cellcolor{secondcolor}\underline{44.8} & 58.1 & \cellcolor{thirdcolor}44.3 & 38.6 & 36.2 & \cellcolor{thirdcolor}15.7 & \cellcolor{secondcolor}\underline{66.1} & 43.4
        & \cellcolor{secondcolor}\underline{85.0} & \cellcolor{secondcolor}\underline{68.0}
        \\
    \midrule
    \textbf{ADRS (ours)}
        & 94.1 & \cellcolor{secondcolor}\underline{75.0} & \cellcolor{topcolor}\textbf{100.0} & \cellcolor{topcolor}\textbf{100.0} & \cellcolor{topcolor}\textbf{90.0} & \cellcolor{topcolor}\textbf{94.7} & \cellcolor{topcolor}\textbf{94.5}
        & \cellcolor{thirdcolor}44.7 & \cellcolor{topcolor}\textbf{61.4} & \cellcolor{topcolor}\textbf{45.8} & \cellcolor{secondcolor}\underline{39.9} & 37.3 & 14.4 & 64.9 & \cellcolor{topcolor}\textbf{45.0}
        & \cellcolor{topcolor}\textbf{87.5} & \cellcolor{topcolor}\textbf{76.6}
        \\

    \midrule
    \rowcolor{gray!10} \multicolumn{18}{l}{\textit{Qwen2.5-7B-Instruct}} \\
    Vanilla
        & 36.1 & 22.2 & 3.1 & 0.0 & 0.0 & 0.0 & 12.5
        & 25.2 & 50.8 & 29.5 & 29.0 & 29.0 & 10.4 & 63.7 & 33.9
        & 5.9 & 1.6
        \\
    Skill-Prompt*
        & 51.7 & 50.0 & 32.3 & 5.3 & 4.3 & 0.0 & 23.4
        & 30.9 & 52.1 & 32.7 & 32.7 & 27.9 & 12.7 & 66.1 & 36.4
        & 1.7 & 0.8
        \\
    OPSD
        & 50.0 & 60.0 & 22.7 & 21.4 & 17.6 & 9.5 & 32.8
        & 8.8 & 8.6 & 17.5 & 2.5 & 4.2 & 0.5 & 1.2 & 6.2
        & 4.5 & 2.3
        \\
    GRPO
        & 91.2 & \cellcolor{secondcolor}\underline{87.5} & 96.2 & 81.0 & 65.0 & 57.9 & 81.2
        & 45.1 & 63.7 & 44.0 & 43.6 & 43.2 & 16.8 & 37.6 & 42.0
        & 80.9 & 72.6
        \\
    Skill-GRPO
        & 88.5 & 66.7 & 65.2 & 61.1 & 57.7 & 73.1 & 69.5
        & 45.2 & 63.7 & 45.7 & 43.1 & 43.3 & 19.6 & 21.4 & 40.3
        & 80.4 & 71.9
        \\
    Skill-GRPO*
        & \cellcolor{topcolor}\textbf{100.0} & 83.3 & 96.4 & 83.3 & 75.0 & \cellcolor{secondcolor}\underline{78.9} & \cellcolor{secondcolor}\underline{88.3}
        & 44.8 & 63.0 & 45.1 & 43.7 & 43.7 & \cellcolor{secondcolor}\underline{20.5} & \cellcolor{thirdcolor}71.4 & 47.5
        & 87.0 & \cellcolor{secondcolor}\underline{81.2}
        \\
    GRPO+OPSD
        & 91.4 & 61.5 & \cellcolor{topcolor}\textbf{100.0} & \cellcolor{thirdcolor}87.5 & \cellcolor{thirdcolor}76.5 & 52.2 & 80.4
        & \cellcolor{secondcolor}\underline{47.3} & \cellcolor{topcolor}\textbf{64.5} & \cellcolor{thirdcolor}46.9 & \cellcolor{thirdcolor}43.8 & 39.3 & 18.0 & 69.4 & 47.0
        & 86.8 & 76.5
        \\
    Skill-SD
        & 93.9 & \cellcolor{topcolor}\textbf{93.8} & 90.9 & \cellcolor{topcolor}\textbf{100.0} & 69.2 & 68.4 & 85.1
        & \cellcolor{thirdcolor}47.1 & \cellcolor{topcolor}\textbf{64.5} & \cellcolor{secondcolor}\underline{47.8} & \cellcolor{secondcolor}\underline{44.2} & 42.1 & \cellcolor{thirdcolor}20.2 & 69.0 & 47.8
        & 86.1 & 76.5
        \\
    RLSD
        & \cellcolor{topcolor}\textbf{100.0} & \cellcolor{secondcolor}\underline{87.5} & 92.3 & 58.8 & \cellcolor{secondcolor}\underline{80.0} & 65.2 & 82.0
        & 46.8 & 63.0 & 44.4 & \cellcolor{topcolor}\textbf{45.5} & \cellcolor{topcolor}\textbf{48.9} & \cellcolor{topcolor}\textbf{21.5} & \cellcolor{topcolor}\textbf{73.0} & \cellcolor{topcolor}\textbf{49.0}
        & \cellcolor{thirdcolor}87.4 & 77.3
        \\
    SDAR
        & 94.7 & 75.0 & \cellcolor{topcolor}\textbf{100.0} & 86.7 & 68.2 & \cellcolor{secondcolor}\underline{78.9} & \cellcolor{thirdcolor}85.9
        & 46.3 & 63.5 & \cellcolor{topcolor}\textbf{48.2} & \cellcolor{thirdcolor}43.8 & \cellcolor{secondcolor}\underline{48.4} & 19.6 & \cellcolor{topcolor}\textbf{73.0} & \cellcolor{topcolor}\textbf{49.0}
        & \cellcolor{topcolor}\textbf{89.4} & \cellcolor{topcolor}\textbf{82.8}
        \\
    \midrule
    \textbf{ADRS (ours)}
        & \cellcolor{topcolor}\textbf{100.0} & \cellcolor{secondcolor}\underline{87.5} & \cellcolor{topcolor}\textbf{100.0} & \cellcolor{secondcolor}\underline{95.2} & \cellcolor{topcolor}\textbf{90.0} & \cellcolor{topcolor}\textbf{94.7} & \cellcolor{topcolor}\textbf{96.1}
        & \cellcolor{topcolor}\textbf{47.5} & \cellcolor{thirdcolor}64.2 & 46.5 & 43.6 & \cellcolor{thirdcolor}43.8 & 18.7 & 69.4 & \cellcolor{thirdcolor}48.2
        & \cellcolor{secondcolor}\underline{88.0} & \cellcolor{thirdcolor}79.7
        \\

    \midrule
    \rowcolor{gray!10} \multicolumn{18}{l}{\textit{Qwen3-1.7B}} \\
    Vanilla
        & 25.0 & 22.2 & 3.1 & 0.0 & 21.4 & 4.2 & 12.5
        & 29.4 & 46.9 & 37.0 & 23.5 & 19.6 & 6.4 & 10.5 & 24.8
        & 46.5 & 4.7
        \\
    Skill-Prompt*
        & 10.3 & \cellcolor{secondcolor}\underline{50.0} & 16.1 & 0.0 & 0.0 & 5.0 & 9.4
        & 29.4 & 46.5 & 36.2 & 22.9 & 20.8 & 4.3 & 10.1 & 24.3
        & 23.0 & 2.3
        \\
    OPSD
        & 26.3 & 33.3 & 9.1 & 0.0 & 4.5 & 5.3 & 14.1
        & 4.2 & 8.3 & 4.6 & 6.6 & 15.3 & 0.7 & 1.2 & 5.8
        & 47.4 & 9.3
        \\
    GRPO
        & \cellcolor{thirdcolor}71.1 & 41.7 & 36.4 & \cellcolor{thirdcolor}40.0 & 31.8 & \cellcolor{thirdcolor}31.6 & 46.1
        & \cellcolor{thirdcolor}40.0 & \cellcolor{secondcolor}\underline{58.9} & 43.5 & 35.4 & 30.3 & 12.0 & \cellcolor{thirdcolor}65.7 & 40.8
        & 67.3 & 38.3
        \\
    Skill-GRPO
        & 27.6 & \cellcolor{topcolor}\textbf{54.5} & 22.7 & 27.3 & 0.0 & 19.2 & 21.1
        & 39.2 & 58.6 & 43.9 & 35.2 & 28.2 & 11.5 & \cellcolor{secondcolor}\underline{66.1} & 40.4
        & 73.4 & 46.1
        \\
    Skill-GRPO*
        & 31.4 & 42.9 & 51.9 & 8.3 & 11.5 & 7.1 & 28.1
        & 38.0 & 58.4 & 43.9 & \cellcolor{secondcolor}\underline{36.3} & 29.0 & \cellcolor{thirdcolor}12.5 & \cellcolor{topcolor}\textbf{66.9} & 40.7
        & \cellcolor{secondcolor}\underline{80.4} & 50.0
        \\
    GRPO+OPSD
        & 38.2 & \cellcolor{secondcolor}\underline{50.0} & 30.8 & 28.6 & 30.0 & 21.1 & 32.0
        & \cellcolor{secondcolor}\underline{40.7} & \cellcolor{secondcolor}\underline{58.9} & 45.0 & \cellcolor{topcolor}\textbf{37.0} & \cellcolor{secondcolor}\underline{34.6} & \cellcolor{topcolor}\textbf{13.3} & \cellcolor{thirdcolor}65.7 & \cellcolor{secondcolor}\underline{42.2}
        & 70.7 & 38.3
        \\
    Skill-SD
        & 52.9 & 37.5 & \cellcolor{secondcolor}\underline{69.2} & \cellcolor{secondcolor}\underline{42.9} & \cellcolor{secondcolor}\underline{60.0} & \cellcolor{topcolor}\textbf{36.8} & \cellcolor{thirdcolor}52.3
        & 39.1 & 57.5 & \cellcolor{secondcolor}\underline{45.4} & 34.8 & \cellcolor{thirdcolor}34.1 & 10.7 & 64.1 & 40.8
        & \cellcolor{topcolor}\textbf{81.8} & \cellcolor{thirdcolor}53.9
        \\
    RLSD
        & 50.0 & 37.5 & \cellcolor{thirdcolor}61.5 & 19.0 & \cellcolor{thirdcolor}50.0 & 21.1 & 42.2
        & 38.6 & 57.3 & 43.0 & 34.5 & \cellcolor{thirdcolor}34.1 & 11.5 & 65.3 & 40.6
        & 74.0 & 50.8
        \\
    SDAR
        & \cellcolor{secondcolor}\underline{73.5} & 25.0 & \cellcolor{topcolor}\textbf{76.9} & 33.3 & 40.0 & \cellcolor{topcolor}\textbf{36.8} & \cellcolor{secondcolor}\underline{53.9}
        & 39.7 & \cellcolor{secondcolor}\underline{58.9} & \cellcolor{thirdcolor}45.3 & \cellcolor{thirdcolor}35.9 & \cellcolor{topcolor}\textbf{35.5} & \cellcolor{secondcolor}\underline{12.6} & 65.3 & \cellcolor{thirdcolor}41.9
        & 76.8 & \cellcolor{secondcolor}\underline{58.6}
        \\
    \midrule
    \textbf{ADRS (ours)}
        & \cellcolor{topcolor}\textbf{86.8} & \cellcolor{secondcolor}\underline{50.0} & 59.1 & \cellcolor{topcolor}\textbf{73.3} & \cellcolor{topcolor}\textbf{63.6} & 15.8 & \cellcolor{topcolor}\textbf{62.5}
        & \cellcolor{topcolor}\textbf{42.7} & \cellcolor{topcolor}\textbf{59.4} & \cellcolor{topcolor}\textbf{47.4} & 35.7 & 32.2 & \cellcolor{thirdcolor}12.5 & \cellcolor{thirdcolor}65.7 & \cellcolor{topcolor}\textbf{42.9}
        & \cellcolor{thirdcolor}77.7 & \cellcolor{topcolor}\textbf{65.6}
        \\
    \bottomrule
    \end{tabular}
    }
\end{table*}

Table~\ref{tab:main_results_no_opid} reports the common 150-step comparison. Three findings organize
the results.

\textbf{\method establishes new state-of-the-art performance across all
three benchmarks with Qwen2.5-3B.}
With Qwen2.5-3B, \method achieves 94.5\% overall success on ALFWorld,
a 45.0\% macro-average on Search-based QA, and a score/success pair of
87.5/76.6 on WebShop. Compared with the strongest baselines, these
results yield improvements of 10.1 percentage points on ALFWorld,
0.4 points on Search-based QA, and 2.5/8.6 points on the two WebShop
metrics, respectively, establishing new state-of-the-art results across
all three benchmarks.

\textbf{\method delivers consistent ALFWorld gains across model families
and scales.}
\method attains the highest ALFWorld overall success in all three model
blocks: 94.5\% with Qwen2.5-3B, 96.1\% with Qwen2.5-7B, and 62.5\%
with Qwen3-1.7B. These results exceed the strongest corresponding
baselines by 10.1, 7.8, and 8.6 percentage points, respectively. With
Qwen3-1.7B, \method also achieves the highest Search-based QA
macro-average of 42.9\% and the highest WebShop success rate of 65.6\%,
improving over the strongest baselines by 0.7 and 7.0 points. These
results demonstrate that the effectiveness of \method extends across
different model families and parameter scales.

\textbf{The performance gains do not rely on privileged information at
evaluation time.}
Skill-Prompt and the starred Skill-GRPO variant provide skills directly
to the policy during evaluation, whereas \method uses only the ordinary
interaction history during rollout and inference. The leading results
of \method therefore do not depend on additional test-time skill
context. Instead, \method converts training-only privileged guidance
into a policy that performs effectively under skill-free interaction.

Appendix Figures~\ref{fig:success-dynamics}--\ref{fig:episode-length-dynamics} shows the training success rate , episode-length, and response-length trajectories. The training curves exhibit consistent upward trends across all configurations.

\subsection{Long-Horizon Training}
\label{sec:convergence}

The 150-step table establishes a common-budget comparison but does not
show whether the gains persist during later optimization. We therefore
continue representative runs to 300 steps and report the best checkpoint,
final checkpoint, and mean over the last three evaluations in
Table~\ref{tab:extended}. These statistics distinguish an isolated peak
from sustained late-training performance.

\textbf{ADRS sustains strong late-training performance on the two
procedural environments.}
On ALFWorld, the ADRS-global+entropy/clip run reaches a 97.7\% peak and
finishes at 96.1\%, 9.4 points above the corresponding GRPO final
checkpoint. This run maintains a high task success rate. Within the
GiGPO sequence, ADRS+TVA improves the final/last-three values from
91.4/93.0\% to 94.5/94.3\%. On WebShop, an ADRS run reaches an 84.4\%
peak and final success rate, 5.5 points above the best GRPO checkpoint;
the no-TVA run records the strongest last-three average at 80.2\%.

\begin{table}[t]
\caption{Representative 300-step convergence results. ``Best'' is the highest observed checkpoint.}
\label{tab:extended}
\centering
\scriptsize
\setlength{\tabcolsep}{3pt}
\begin{tabular}{llrrr}
\toprule
Env. & Configuration & Best & Final & Last-3 \\
\midrule
\multirow{4}{*}{ALF.}
 & GRPO & 92.2 & 86.7 & 86.2 \\
 & GiGPO & 96.9 & 91.4 & 93.0 \\
 & ADRS-global (GRPO) & \textbf{97.7} & \textbf{96.1} & \textbf{96.1} \\
 & ADRS+TVA (GiGPO) & 96.9 & 94.5 & 94.3 \\
\midrule
\multirow{4}{*}{Web.}
 & GRPO & 78.9 & 78.9 & 75.3 \\
 & ADRS (no TVA) & 82.8 & 82.8 & \textbf{80.2} \\
 & ADRS-per-seq & \textbf{84.4} & \textbf{84.4} & 77.9 \\
 & ADRS-per-seq+GateNorm & \textbf{84.4} & 82.0 & 78.9 \\
\bottomrule
\end{tabular}
\end{table}

\begin{figure}[t]
    \centering
    \includegraphics[width=\columnwidth]{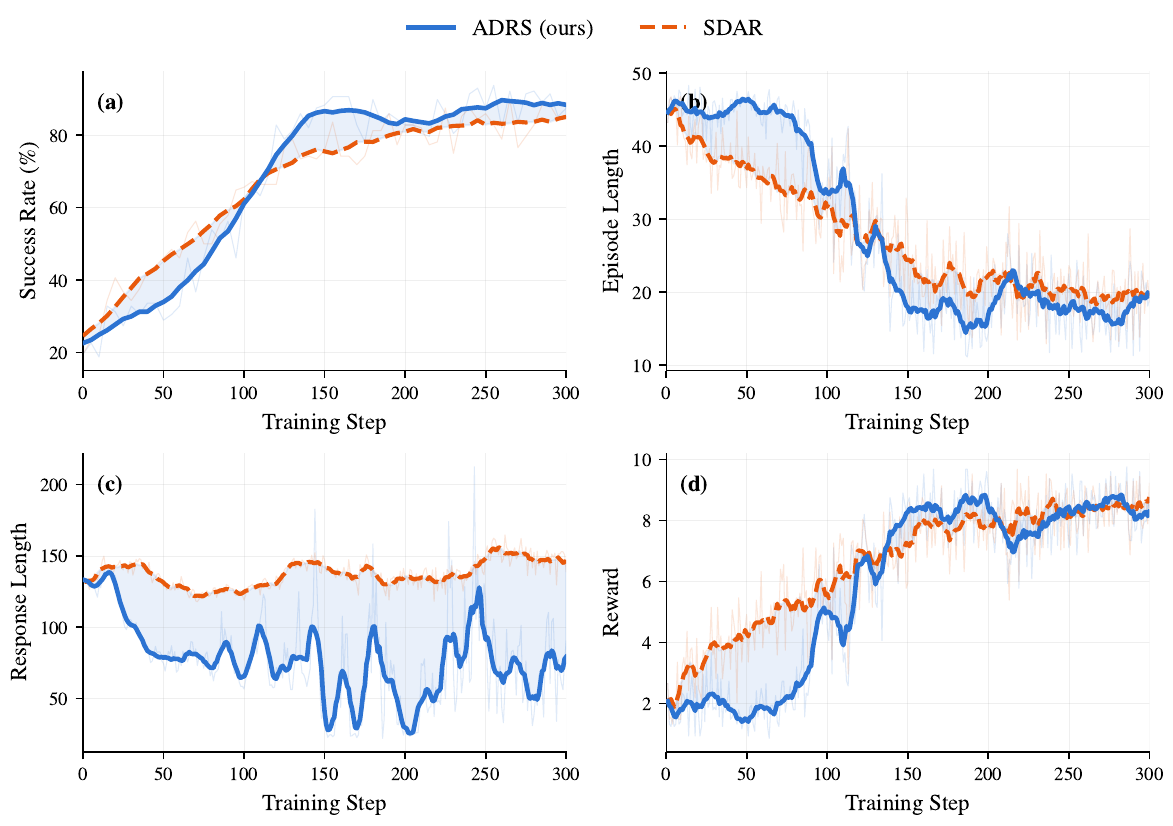}
    \vspace{-20pt}
    \Description{Training curves on ALFWorld through 300 policy updates, comparing SDAR with the ADRS+TVA run on the GiGPO backbone. Light lines show checkpoint traces and dark lines show smoothed trends.}
    \caption{\textbf{ALFWorld training dynamics through 300 steps.}
    Curves compare SDAR and ADRS+TVA (GiGPO).}
    \label{fig:alfworld-300}
\end{figure}

Figure~\ref{fig:alfworld-300} visualizes the 300-step ALFWorld
trajectories. Appendix Table~\ref{tab:extended-full} reports the complete
named-run matrix.

\subsection{Ablations on GRPO and GiGPO Backbones}
\label{sec:ablation}

\begin{figure}[t]
    \centering
    \includegraphics[width=\columnwidth]{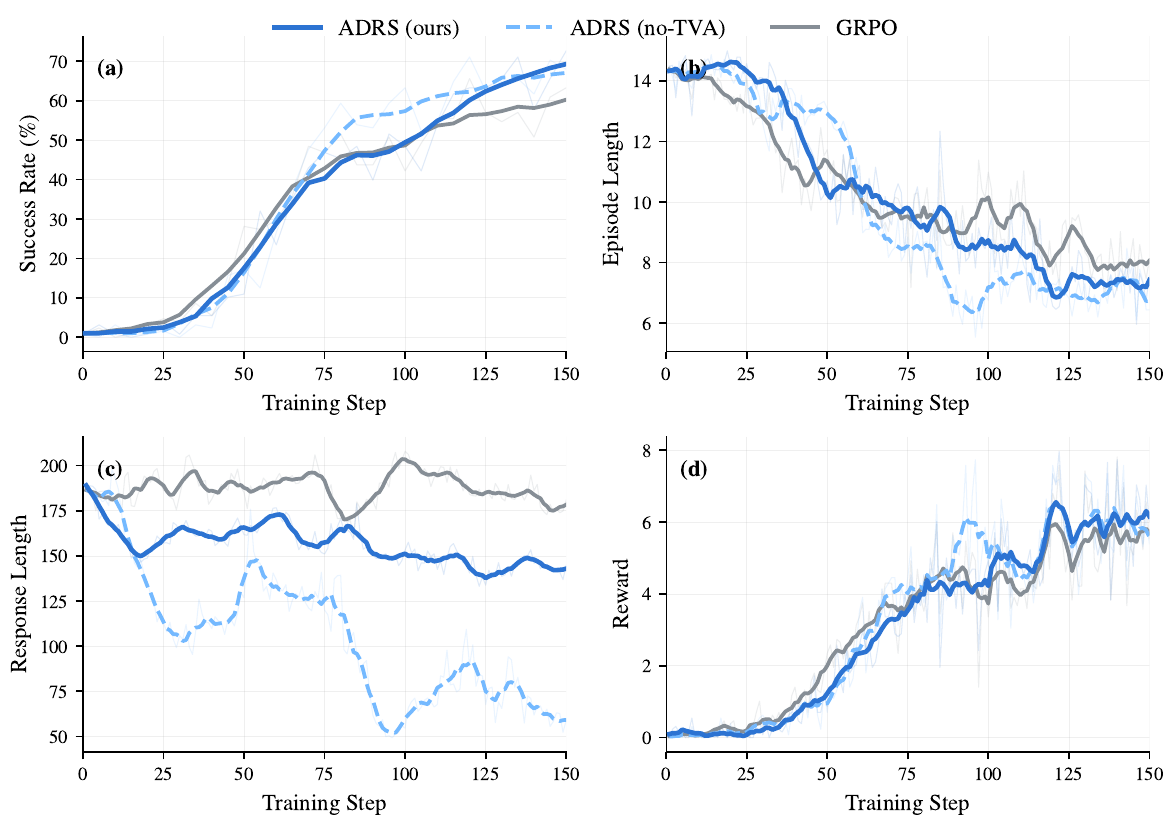}
    \vspace{-20pt}
    \Description{Training curves on WebShop through 150 policy updates, comparing GRPO with the corresponding ADRS variants.}
    \caption{\textbf{WebShop ablation dynamics through 150 steps.}
    Curves compare GRPO with the ADRS variants.}
    \label{fig:webshop-150}
\end{figure}

\begin{figure}[t]
    \centering
    \includegraphics[width=\columnwidth]{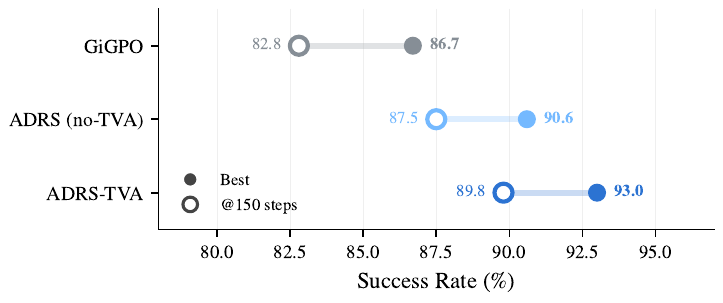}
    \vspace{-20pt}
    \Description{An ALFWorld ablation chart comparing GiGPO, ADRS without TVA, and ADRS with TVA by step-150 and best-checkpoint success rates.}
    \caption{GiGPO-based ALFWorld ablation on Qwen2.5-3B. The chart
    compares GiGPO, ADRS without TVA, and ADRS with TVA at step 150 and
    at their best observed checkpoints.}
    \label{fig:backbone-ablation}
\end{figure}

Figure~\ref{fig:webshop-150} shows the WebShop training dynamics under
the 150-step budget, comparing GRPO with the corresponding ADRS variants.
The ADRS variants remain above GRPO over most of training, and the
strongest configuration reaches higher final performance at step 150.
This complements the endpoint comparison in the main table, showing that
the WebShop gain is visible across the training curve rather than only at
a single checkpoint.

Figure~\ref{fig:backbone-ablation} compares GiGPO, ADRS without TVA, and
ADRS+TVA on ALFWorld, testing whether ADRS still brings gains when the
backbone already provides step-aware credit. The WebShop trajectories
show that ADRS improves a GRPO backbone under the preferred WebShop
normalization during optimization. Appendix Tables~\ref{tab:normalization} and~\ref{tab:eta}
report the matched normalization comparison and teacher-scale sensitivity
results.

\textbf{ADRS remains beneficial on the GiGPO backbone.}
At step 150, centered ADRS improves GiGPO from 82.8\% to 87.5\%, and
ADRS+TVA further reaches 89.8\%. Their corresponding best checkpoints
are 86.7\%, 90.6\%, and 93.0\%. Across the 300-step runs, TVA increases
the peak by 0.8 points and the last-three average by 2.4 points. This
comparison supports the complete GiGPO-based system. Appendix
Table~\ref{tab:ablation-full} preserves the fully named matrix.

\subsection{Action--Object Mechanism Diagnostic}
\label{sec:mechanism-diagnostic}

\begin{figure}[t]
    \centering
    \includegraphics[width=\columnwidth]{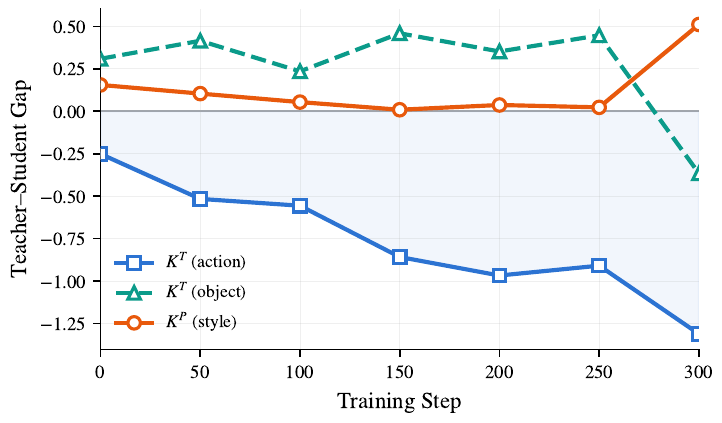}
    \vspace{-25pt}
    \Description{Line chart from training step 0 to 300 showing signed teacher-student log-probability gaps for action, object, and style tokens. The action gap becomes increasingly negative, the object gap remains positive until step 250 and reverses at step 300, and the style gap stays near zero until rising at step 300.}
    \caption{Action--object diagnostic over training on four fixed
    successful ALFWorld trajectories. $K^T$ groups task-bearing action
    and object tokens, whereas $K^P$ denotes the residual style-token
    group. Values are signed teacher--student log-probability gaps.}
    \label{fig:kt-kp-diagnostic}
\end{figure}

To examine how token-level advantage modulation supports long-horizon
performance, we probe four fixed successful ALFWorld trajectories: heat,
pick-and-place, cool, and look-at-object-in-light. These trajectories
contain 32 action-token, 80 object-token, and 470 style/other-token
observations. Figure~\ref{fig:kt-kp-diagnostic} groups the signed
teacher--student gaps into $K^T$ action/object terms and a residual
$K^P$ style term, revealing how privileged rescoring separates
task-bearing tokens from generic style tokens over training.

At step 150, the action gap reaches $-0.8582$ and the object gap remains
positive at $0.4591$, while the style gap contracts to $0.0085$, yielding
an action-to-style magnitude ratio of 95.3. This separation shows that
the token-level ADRS signal increasingly concentrates update contrast on
tokens that determine executable actions and object choices, rather than
spreading credit uniformly across the generated response. The pattern
explains why token-level advantage modulation improves long-horizon
interaction: it strengthens learning on the local decisions that carry
delayed task consequences. After step 250, task success approaches
saturation and the policy changes more slowly; the shift in the object
term and the smaller final ratio coincide with this later training stage.

\subsection{Data Efficiency and Unseen Generalization}
\label{sec:data-generalization}

\begin{figure}[t]
    \centering
    \begin{minipage}[t]{0.48\linewidth}
        \centering
        \includegraphics[width=\linewidth]{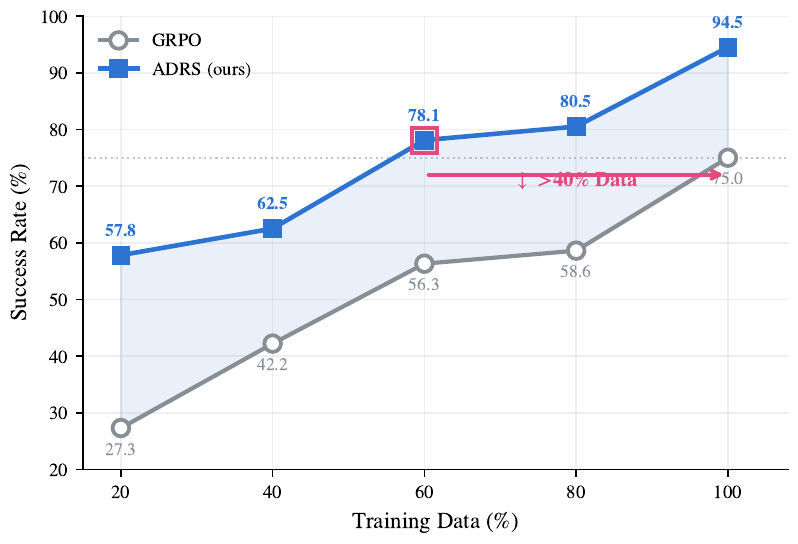}
    \end{minipage}\hfill
    \begin{minipage}[t]{0.48\linewidth}
        \centering
        \includegraphics[width=\linewidth]{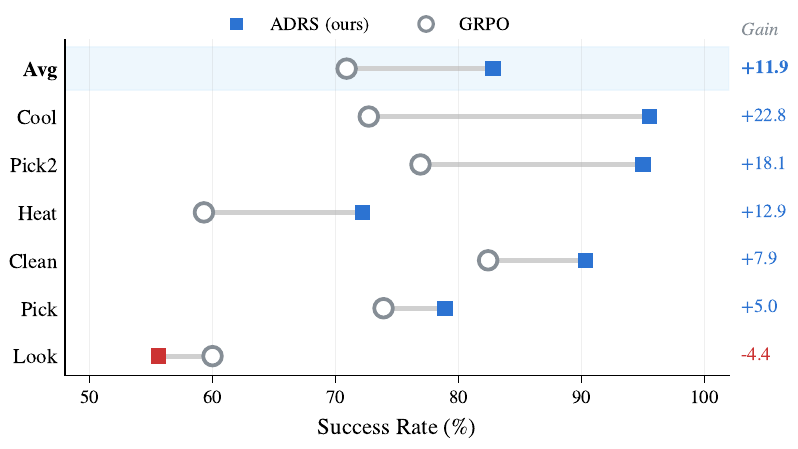}
    \end{minipage}
    \vspace{-5pt}
    \Description{The left chart plots ADRS ALFWorld success with 20, 40, 60, 80, and 100 percent of the training data and marks full-data GRPO at 75.0 percent success. The right chart compares ADRS and GRPO on six unseen ALFWorld task categories and their average, with differences printed for each row.}
    \caption{Data efficiency and evaluation on the supplied ALFWorld
    unseen split. Left: ADRS trained on five fractions of the available
    data, with full-data GRPO shown as a reference. Right: category-level
    success on the unseen evaluation.}
    \label{fig:data-generalization}
\end{figure}

We next evaluate whether ADRS can extract useful token-level credit from
smaller training sets and whether the learned behavior transfers to the
supplied unseen ALFWorld split.

\textbf{ADRS surpasses full-data GRPO after using 60\% of the training
data.}
Figure~\ref{fig:data-generalization} (left) evaluates ADRS with 20\%,
40\%, 60\%, 80\%, and 100\% of the available ALFWorld training data.
ADRS obtains 57.8\%, 62.5\%, 78.1\%, 80.5\%, and 94.5\% success,
respectively. The 60\%, 80\%, and 100\% settings exceed the 75.0\%
full-data GRPO reference, while the 20\% and 40\% settings remain below
it. This trend shows that ADRS benefits from additional training data
but can exceed the full-data outcome-only baseline before using the
entire training set.

\textbf{ADRS transfers to most categories in the supplied unseen split.}
Figure~\ref{fig:data-generalization} (right) compares ADRS and GRPO on
the unseen ALFWorld split. ADRS improves the average success rate by
11.9 points. The category-level differences are +22.8 on Cool, +18.1 on
Pick2, +12.9 on Heat, +7.9 on Clean, +5.0 on Pick, and $-4.4$ on Look.
Thus, ADRS improves five of the six reported categories, with the largest
gains on state-changing and multi-object tasks.

\subsection{Effect of Maximum Interaction Steps}
\label{sec:horizon-sweep}

\begin{figure}[t]
    \centering
    \includegraphics[width=\columnwidth]{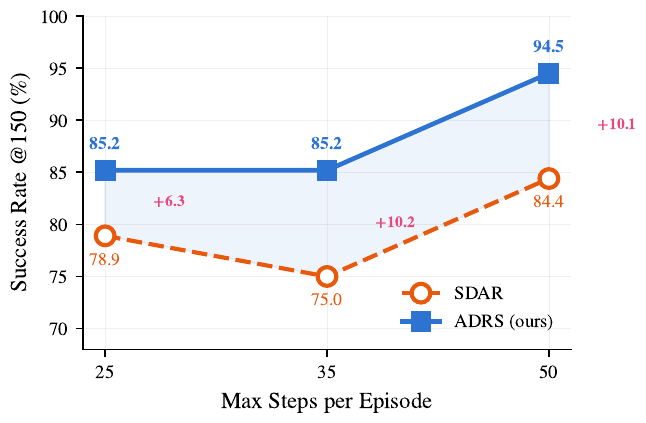}
    \vspace{-25pt}
    \Description{Grouped bar chart of ALFWorld success at training step 150 for SDAR and ADRS with maximum episode lengths of 25, 35, and 50 steps. ADRS outperforms SDAR across the three settings.}
    \caption{Effect of maximum interaction steps on ALFWorld with
    Qwen2.5-3B at training step 150.}
    \label{fig:horizon-sweep}
\end{figure}

Figure~\ref{fig:horizon-sweep} compares ADRS and SDAR with maximum episode lengths of 25, 35, and 50 steps. ADRS consistently outperforms SDAR, achieving 85.2\%, 85.2\%, and 94.5\%, respectively, compared with 78.9\%, 75.0\%, and 84.4\% for SDAR. As the interaction budget increases, ADRS maintains its advantage and reaches its strongest result at the 50-step horizon. This result indicates that ADRS can use a longer interaction horizon to improve long-horizon task performance while preserving reliable token-level credit assignment across interaction steps. 



\section{Conclusion}


We presented \method, a self-distilled reward-shaping framework for sparse temporal credit in multi-turn language-agent reinforcement learning. \method transforms task-matched procedural guidance into token rewards through within-step calibration, return-associated \tva modulation, and pre-advantage integration. These components determine what the privileged teacher prefers, when that preference is return-relevant, and how it enters policy optimization, while keeping rollouts and inference skill-free. Experiments across three interactive benchmarks and multiple training settings demonstrate consistent improvements, particularly on long-horizon tasks. Taken together, these results indicate that training-only procedural knowledge can be converted into return-associated token-level credit and incorporated directly into the native reward-to-advantage construction.
\clearpage
\bibliographystyle{ACM-Reference-Format}
\bibliography{references}

@article{lu2026self,
  title={Self-distilled agentic reinforcement learning},
  author={Lu, Zhengxi and Yao, Zhiyuan and Han, Zhuowen and Wang, Zi-Han and Wu, Jinyang and Gu, Qi and Cai, Xunliang and Lu, Weiming and Xiao, Jun and Zhuang, Yueting and others},
  journal={arXiv preprint arXiv:2605.15155},
  year={2026}
}

@article{zhao2026self,
  title={Self-Distilled Reasoner: On-Policy Self-Distillation for Large Language Models},
  author={Zhao, Siyan and Xie, Zhihui and Liu, Mengchen and Huang, Jing and Pang, Guan and Chen, Feiyu and Grover, Aditya},
  journal={arXiv preprint arXiv:2601.18734},
  year={2026}
}

@article{feng2026group,
  title={{Group-in-group policy optimization for {LLM} agent training}},
  author={Feng, Lang and Xue, Zhenghai and Liu, Tingcong and An, Bo},
  journal={Advances in Neural Information Processing Systems},
  volume={38},
  pages={46375--46408},
  year={2026}
}

@article{jin2025search,
  title={{Search-R1}: Training {LLM}s to Reason and Leverage Search Engines with Reinforcement Learning},
  author={Jin, Bowen and Zeng, Hansi and Yue, Zhenrui and Yoon, Jinsung and Arik, Sercan and Wang, Dong and Zamani, Hamed and Han, Jiawei},
  journal={arXiv preprint arXiv:2503.09516},
  year={2025}
}

@article{shao2024deepseekmath,
  title={{DeepSeekMath}: Pushing the limits of mathematical reasoning in open language models},
  author={Shao, Zhihong and Wang, Peiyi and Zhu, Qihao and Xu, Runxin and Song, Junxiao and Bi, Xiao and Zhang, Haowei and Zhang, Mingchuan and Li, YK and Wu, Yang and others},
  journal={arXiv preprint arXiv:2402.03300},
  year={2024}
}

@article{shridhar2020alfworld,
  title={{ALFWorld}: Aligning text and embodied environments for interactive learning},
  author={Shridhar, Mohit and Yuan, Xingdi and C{\^o}t{\'e}, Marc-Alexandre and Bisk, Yonatan and Trischler, Adam and Hausknecht, Matthew},
  journal={arXiv preprint arXiv:2010.03768},
  year={2020}
}

@article{yao2022webshop,
  title={{WebShop}: Towards scalable real-world web interaction with grounded language agents},
  author={Yao, Shunyu and Chen, Howard and Yang, John and Narasimhan, Karthik},
  journal={Advances in Neural Information Processing Systems},
  volume={35},
  pages={20744--20757},
  year={2022}
}

@misc{qwen2025qwen25technicalreport,
      title={Qwen2.5 Technical Report}, 
      author={{Qwen Team}},
      year={2025},
      eprint={2412.15115},
      archivePrefix={arXiv},
      primaryClass={cs.CL},
      url={https://arxiv.org/abs/2412.15115}, 
}

@article{yang2025qwen3,
  title={Qwen3 technical report},
  author={Yang, An and Li, Anfeng and Yang, Baosong and Zhang, Beichen and Hui, Binyuan and Zheng, Bo and Yu, Bowen and Gao, Chang and Huang, Chengen and Lv, Chenxu and others},
  journal={arXiv preprint arXiv:2505.09388},
  year={2025}
}

@article{schulman2017proximal,
  title={Proximal policy optimization algorithms},
  author={Schulman, John and Wolski, Filip and Dhariwal, Prafulla and Radford, Alec and Klimov, Oleg},
  journal={arXiv preprint arXiv:1707.06347},
  year={2017}
}

@inproceedings{ng1999policy,
  title={Policy invariance under reward transformations: Theory and application to reward shaping},
  author={Ng, Andrew Y and Harada, Daishi and Russell, Stuart},
  booktitle={Icml},
  volume={99},
  pages={278--287},
  year={1999},
  organization={Citeseer}
}

@article{vapnik2009new,
  title={A new learning paradigm: Learning using privileged information},
  author={Vapnik, Vladimir and Vashist, Akshay},
  journal={Neural networks},
  volume={22},
  number={5-6},
  pages={544--557},
  year={2009},
  publisher={Elsevier}
}

@article{hinton2015distilling,
  title={Distilling the knowledge in a neural network},
  author={Hinton, Geoffrey and Vinyals, Oriol and Dean, Jeff},
  journal={arXiv preprint arXiv:1503.02531},
  year={2015}
}

@article{wang2026skill,
  title={{Skill-SD}: Skill-Conditioned Self-Distillation for Multi-Turn {LLM} Agents},
  author={Wang, Hao and Wang, Guozhi and Xiao, Han and Zhou, Yufeng and Pan, Yue and Wang, Jichao and Xu, Ke and Wen, Yafei and Ruan, Xiaohu and Chen, Xiaoxin and others},
  journal={arXiv preprint arXiv:2604.10674},
  year={2026}
}

@article{yang2026self,
  title={Self-Distilled {RLVR}},
  author={Yang, Chenxu and Qin, Chuanyu and Si, Qingyi and Chen, Minghui and Gu, Naibin and Yao, Dingyu and Lin, Zheng and Wang, Weiping and Wang, Jiaqi and Duan, Nan},
  journal={arXiv preprint arXiv:2604.03128},
  year={2026}
}

@article{luo2025agent,
  title={{Agent Lightning}: Train Any {AI} Agents with Reinforcement Learning},
  author={Luo, Xufang and Zhang, Yuge and He, Zhiyuan and Wang, Zilong and Zhao, Siyun and Li, Dongsheng and Qiu, Luna K and Yang, Yuqing},
  journal={arXiv preprint arXiv:2508.03680},
  year={2025}
}

@article{liu2026self,
  title={Self-Distilled Policy Gradient},
  author={Liu, Yifeng and Zhang, Shiyuan and Zhang, Yifan and Gu, Quanquan},
  journal={arXiv preprint arXiv:2606.04036},
  year={2026}
}

@article{kim2026rebellious,
  title={{Rebellious Student}: Reversing Teacher Signals for Reasoning Exploration with Self-Distilled {RLVR}},
  author={Kim, Jeonghye and Jeon, Jiwon and Li, Dongsheng and Yang, Yuqing},
  journal={arXiv preprint arXiv:2605.10781},
  year={2026}
}

@article{meng2026craft,
  title={CRAFT: Counterfactual Credit Assignment from Free Sibling Rollouts for Self-Distilled Agentic Reinforcement Learning},
  author={Meng, Zibin and Chen, Kani},
  journal={arXiv preprint arXiv:2606.29476},
  year={2026}
}

@article{arjona2019rudder,
  title={Rudder: Return decomposition for delayed rewards},
  author={Arjona-Medina, Jose A and Gillhofer, Michael and Widrich, Michael and Unterthiner, Thomas and Brandstetter, Johannes and Hochreiter, Sepp},
  journal={Advances in Neural Information Processing Systems},
  volume={32},
  year={2019}
}

@inproceedings{agarwal2024policy,
  title={On-policy distillation of language models: Learning from self-generated mistakes},
  author={Agarwal, Rishabh and Vieillard, Nino and Zhou, Yongchao and Stanczyk, Piotr and Ramos Garea, Sabela and Geist, Matthieu and Bachem, Olivier},
  booktitle={International Conference on Learning Representations},
  volume={2024},
  pages={21246--21263},
  year={2024}
}

@inproceedings{ross2011reduction,
  title={A reduction of imitation learning and structured prediction to no-regret online learning},
  author={Ross, St{\'e}phane and Gordon, Geoffrey and Bagnell, Drew},
  booktitle={Proceedings of the fourteenth international conference on artificial intelligence and statistics},
  pages={627--635},
  year={2011},
  organization={JMLR Workshop and Conference Proceedings}
}

@article{snell2022learning,
  title={Learning by distilling context},
  author={Snell, Charlie and Klein, Dan and Zhong, Ruiqi},
  journal={arXiv preprint arXiv:2209.15189},
  year={2022}
}

@article{hubotter2026reinforcement,
  title={Reinforcement Learning via Self-Distillation},
  author={H{\"u}botter, Jonas and L{\"u}beck, Frederike and Behric, Lejs and Baumann, Anton and Bagatella, Marco and Marta, Daniel and Hakimi, Ido and Shenfeld, Idan and Buening, Thomas Kleine and Guestrin, Carlos and others},
  journal={arXiv preprint arXiv:2601.20802},
  year={2026}
}

@article{xu2025kdrl,
  title={{KDRL}: Post-Training Reasoning {LLM}s via Unified Knowledge Distillation and Reinforcement Learning},
  author={Xu, Hongling and Zhu, Qi and Deng, Heyuan and Li, Jinpeng and Hou, Lu and Wang, Yasheng and Shang, Lifeng and Xu, Ruifeng and Mi, Fei},
  journal={arXiv preprint arXiv:2506.02208},
  year={2025}
}

@article{wang2026tcod,
  title={{TCOD}: Exploring Temporal Curriculum in On-Policy Distillation for Multi-Turn Autonomous Agents},
  author={Wang, Jiaqi and Zhang, Wenhao and Shi, Weijie and Li, Yaliang and Cheng, James},
  journal={arXiv preprint arXiv:2604.24005},
  year={2026}
}

@article{ding2026hdpo,
  title={{HDPO}: Hybrid Distillation Policy Optimization via Privileged Self-Distillation},
  author={Ding, Ken},
  journal={arXiv preprint arXiv:2603.23871},
  year={2026}
}

@article{xu2026tip,
  title={{TIP}: Token Importance in On-Policy Distillation},
  author={Xu, Yuanda and Sang, Hejian and Zhou, Zhengze and He, Ran and Wang, Zhipeng and Geramifard, Alborz},
  journal={arXiv preprint arXiv:2604.14084},
  year={2026}
}

@article{ye2026policy,
  title={On-policy context distillation for language models},
  author={Ye, Tianzhu and Dong, Li and Wu, Xun and Huang, Shaohan and Wei, Furu},
  journal={arXiv preprint arXiv:2602.12275},
  year={2026}
}

@article{chen2025reinforcement,
  title={Reinforcement Learning for Long-Horizon Interactive {LLM} Agents},
  author={Chen, Kevin and Cusumano-Towner, Marco and Huval, Brody and Petrenko, Aleksei and Hamburger, Jackson and Koltun, Vladlen and Kr{\"a}henb{\"u}hl, Philipp},
  journal={arXiv preprint arXiv:2502.01600},
  year={2025}
}

@article{wang2025ragen,
  title={{RAGEN}: Understanding Self-Evolution in {LLM} Agents via Multi-Turn Reinforcement Learning},
  author={Wang, Zihan and Wang, Kangrui and Wang, Qineng and Zhang, Pingyue and Li, Linjie and Yang, Zhengyuan and Jin, Xing and Yu, Kefan and Nguyen, Minh Nhat and Liu, Licheng and others},
  journal={arXiv preprint arXiv:2504.20073},
  year={2025}
}

@inproceedings{lightman2024let,
  title={Let's verify step by step},
  author={Lightman, Hunter and Kosaraju, Vineet and Burda, Yuri and Edwards, Harrison and Baker, Bowen and Lee, Teddy and Leike, Jan and Schulman, John and Sutskever, Ilya and Cobbe, Karl},
  booktitle={International Conference on Learning Representations},
  volume={2024},
  pages={39578--39601},
  year={2024}
}

@article{xia2026skillrl,
  title={{SkillRL}: Evolving Agents via Recursive Skill-Augmented Reinforcement Learning},
  author={Xia, Peng and Chen, Jianwen and Wang, Hanyang and Liu, Jiaqi and Zeng, Kaide and Wang, Yu and Han, Siwei and Zhou, Yiyang and Zhao, Xujiang and Chen, Haifeng and others},
  journal={arXiv preprint arXiv:2602.08234},
  year={2026}
}

@article{kwiatkowski2019natural,
  title={Natural questions: a benchmark for question answering research},
  author={Kwiatkowski, Tom and Palomaki, Jennimaria and Redfield, Olivia and Collins, Michael and Parikh, Ankur and Alberti, Chris and Epstein, Danielle and Polosukhin, Illia and Devlin, Jacob and Lee, Kenton and others},
  journal={Transactions of the Association for Computational Linguistics},
  volume={7},
  pages={453--466},
  year={2019},
  publisher={MIT Press One Rogers Street, Cambridge, MA 02142-1209, USA journals-info~…}
}

@inproceedings{joshi2017triviaqa,
  title={Triviaqa: A large scale distantly supervised challenge dataset for reading comprehension},
  author={Joshi, Mandar and Choi, Eunsol and Weld, Daniel S and Zettlemoyer, Luke},
  booktitle={Proceedings of the 55th Annual Meeting of the Association for Computational Linguistics (Volume 1: Long Papers)},
  pages={1601--1611},
  year={2017}
}

@inproceedings{mallen2023trust,
  title={When not to trust language models: Investigating effectiveness of parametric and non-parametric memories},
  author={Mallen, Alex and Asai, Akari and Zhong, Victor and Das, Rajarshi and Khashabi, Daniel and Hajishirzi, Hannaneh},
  booktitle={Proceedings of the 61st annual meeting of the association for computational linguistics (volume 1: Long papers)},
  pages={9802--9822},
  year={2023}
}

@inproceedings{yang2018hotpotqa,
  title={HotpotQA: A dataset for diverse, explainable multi-hop question answering},
  author={Yang, Zhilin and Qi, Peng and Zhang, Saizheng and Bengio, Yoshua and Cohen, William and Salakhutdinov, Ruslan and Manning, Christopher D},
  booktitle={Proceedings of the 2018 conference on empirical methods in natural language processing},
  pages={2369--2380},
  year={2018}
}

@inproceedings{ho2020constructing,
  title={Constructing a multi-hop qa dataset for comprehensive evaluation of reasoning steps},
  author={Ho, Xanh and Nguyen, Anh-Khoa Duong and Sugawara, Saku and Aizawa, Akiko},
  booktitle={Proceedings of the 28th International Conference on Computational Linguistics},
  pages={6609--6625},
  year={2020}
}

@article{trivedi2022musique,
  title={{MuSiQue}: Multihop Questions via Single-Hop Question Composition},
  author={Trivedi, Harsh and Balasubramanian, Niranjan and Khot, Tushar and Sabharwal, Ashish},
  journal={Transactions of the Association for Computational Linguistics},
  volume={10},
  pages={539--554},
  year={2022},
  publisher={MIT Press One Broadway, 12th Floor, Cambridge, Massachusetts 02142, USA~…}
}

@inproceedings{press2023measuring,
  title={Measuring and narrowing the compositionality gap in language models},
  author={Press, Ofir and Zhang, Muru and Min, Sewon and Schmidt, Ludwig and Smith, Noah A and Lewis, Mike},
  booktitle={Findings of the Association for Computational Linguistics: EMNLP 2023},
  pages={5687--5711},
  year={2023}
}

\clearpage
\appendix
\section{Experimental and Implementation Details}
\label{app:implementation}

This section expands the compact setup in Section~\ref{sec:setup}. It describes the benchmark protocols and baseline groups, then documents the skill provider, named configurations, and optimization settings used in our experiments.

\subsection{Benchmarks}
\label{app:benchmarks}

Our evaluation covers embodied household control, interactive web navigation, and search-augmented question answering. Table~\ref{tab:benchmark-details} summarizes the evaluation sets, primary metrics, and maximum interaction horizons.

\begin{table*}[!t]
\caption{Benchmark and evaluation details for the three agent environments.}
\label{tab:benchmark-details}
\centering
\small
\begin{tabular}{lllll}
\toprule
Benchmark & Interaction & Development / evaluation use & Primary metric & Max turns \\
\midrule
ALFWorld & Text household control & 128 validation tasks & Success (\%) & 50 \\
WebShop & Simulated web shopping & 128 tasks/checkpoint from goals 0--499 & Score and success (\%) & 15 \\
Search & Search-augmented QA & 504-example development; 51K validation & Seven-dataset macro EM (\%) & 4 \\
\bottomrule
\end{tabular}
\end{table*}

\paragraph{ALFWorld.}
ALFWorld~\cite{shridhar2020alfworld} presents household tasks through a text interface derived from embodied environments. Given an instruction and textual observations, the agent must select admissible actions while satisfying object-state and receptacle preconditions. We report success over 128 validation tasks spanning Pick, Look, Clean, Heat, Cool, and Pick2. The 50-turn limit makes ALFWorld the longest-horizon benchmark in the primary suite and exposes errors in ordering, state tracking, and recovery from invalid actions.

\paragraph{WebShop.}
WebShop~\cite{yao2022webshop} requires an agent to search a simulated product catalog, inspect candidate attributes, preserve constraints from a natural-language request, and complete a purchase. Each checkpoint evaluates 128 tasks sampled from goals 0--499. The environment reports a normalized task score for partial constraint satisfaction and a binary success measure for exact completion. Its 15-turn interaction budget is shorter than ALFWorld but still requires multi-stage search, comparison, and selection.

\paragraph{Search-based QA}
Following Search-R1~\cite{jin2025search}, the agent issues search queries, reads retrieved evidence, and returns an answer within four tool-use turns. The aggregate covers Natural Questions~\cite{kwiatkowski2019natural}, TriviaQA~\cite{joshi2017triviaqa}, PopQA~\cite{mallen2023trust},
HotpotQA~\cite{yang2018hotpotqa}, 2WikiMultihopQA~\cite{ho2020constructing}, MuSiQue~\cite{trivedi2022musique}, and Bamboogle~\cite{press2023measuring}. These datasets mix direct retrieval and multi-hop evidence synthesis. The primary runs evaluate the full seven-dataset 51K collection every 75 optimizer steps, whereas the $\eta$ sweep uses a 504-example development subset every 15 steps.

\subsection{Metrics and Reward Construction}

For $N$ evaluation tasks and seven Search datasets $\mathcal D$, the reported aggregates are
\begin{align*}
\mathrm{Succ}_{\mathrm{ALF}}
&=\frac{100}{N}\sum_{n=1}^{N}\mathbf 1[\text{task }n\text{ succeeds}],\\
\mathrm{EM}_{\mathrm{Search}}
&=\frac{1}{|\mathcal D|}\sum_{d\in\mathcal D}
\frac{100}{N_d}\sum_{n=1}^{N_d}\mathbf 1[\widehat a_{d,n}=a_{d,n}],\\
\mathrm{Score}_{\mathrm{WS}}
&=\frac{100}{N}\sum_{n=1}^{N}\mathrm{taskscore}_n,\\
\mathrm{Succ}_{\mathrm{WS}}
&=\frac{100}{N}\sum_{n=1}^{N}\mathbf 1[\mathrm{taskscore}_n=1].
\end{align*}
ALFWorld also reports success by task family, and Search reports the exact match for each constituent dataset before taking their unweighted macro-average. WebShop score and success are complementary: the former preserves partial task completion, whereas the latter requires every purchase constraint to be satisfied. All evaluations use temperature-0.4 sampling.

The episode reward manager places accumulated outcome reward on the final valid response token before advantage estimation. ALFWorld uses $10\,\mathbf 1[\mathrm{won}]$, WebShop uses $10\,\mathbf 1[\mathrm{done}\wedge\mathrm{taskscore}=1]$, and Search uses $\mathbf 1[\mathrm{exact\ match}]$. Invalid-action penalty coefficients are 0.1 for ALFWorld/WebShop and 0.01 for Search; reference-policy coefficients are 0.01, 0.01, and 0.001, respectively. These terms form $r^{\mathrm{base}}$ before Eq.~\ref{eq:star-reward}.

\subsection{Baselines}
\label{app:baselines}

The baselines differ in whether they optimize only terminal outcomes, expose natural-language skills to the acting policy, or use a privileged teacher to supply token-level supervision.

\paragraph{Prompting and outcome-only reinforcement learning.}
Vanilla evaluates the original instruction-tuned backbone without post-training. Skill-Prompt uses the same frozen policy but appends a task-relevant skill during evaluation, isolating the benefit of direct inference-time guidance. GRPO~\cite{shao2024deepseekmath} samples multiple trajectories per task and derives a group-relative advantage from scalar outcomes. Skill-GRPO retains the GRPO objective while supplying skills during training, and the starred variant also supplies them during evaluation. GiGPO~\cite{feng2026group} augments outcome-driven learning with step-aware group credit and serves as a stronger backbone in the 150-step component study.

\paragraph{Privileged self-distillation.}
OPSD~\cite{zhao2026self} uses ordinary and privileged contexts to obtain student and teacher predictions from the same model. GRPO+OPSD combines its auxiliary token objective with outcome-based RL. Skill-SD~\cite{wang2026skill} supplies trajectory-derived skills to the teacher branch, while RLSD~\cite{yang2026self} uses teacher--student differences to modulate token updates. SDAR~\cite{lu2026self} introduces retrieved procedural skills and a confidence-gated auxiliary distillation loss. These methods provide the closest comparison for determining whether privileged guidance is more effective when injected into the native reward-to-advantage path.

\paragraph{Comparison setup.}
All unstarred methods are evaluated without privileged skills. We follow the benchmark, training, and evaluation setup of SDAR~\cite{lu2026self} and use the same interaction protocols across the compared model configurations.

\subsection{Skill Provider and Memory Construction}

The provider contains 37 general skills and 90 task-specific skills. All 619 source memories come from the designated training splits: 223 for ALFWorld (113 successful and 110 failed), 196 for Search (132/64), and 200 for WebShop (120/80). The skill configuration follows the official SDAR setup~\cite{lu2026self}, which adopts the SkillBank introduced by SkillRL~\cite{xia2026skillrl}. The primary runs use compact skill documents rather than appending individual memory records to the teacher prompt.

Selection is deterministic rather than nearest-neighbor retrieval. The provider first selects an environment-level guide and then appends one task-matched file. ALFWorld maps six household task types to five skill files; Search maps metadata to direct-retrieval, multi-hop, entity-attribute, or comparison guidance; WebShop maps the requested product to one of seven categories. Skill text shares the prompt budget (2,048 tokens for ALFWorld and 4,096 for WebShop/Search). The ordinary branch and inference-time policy never receive this text.

The teacher prefix is inserted verbatim as \texttt{[Privileged Skill Information]}, followed by a newline, the selected skill, and two newlines before the ordinary environment prompt. ALFWorld and WebShop require reasoning inside \texttt{<think>} tags and one admissible action inside \texttt{<action>}; Search alternates \texttt{<search>} queries with \texttt{<information>} observations and emits either one search query or a final \texttt{<answer>} per turn.

\subsection{Optimization and Named Configurations}

Tables~\ref{tab:common-setup} and~\ref{tab:setup} collect the archived training configuration.

\begin{table}[!ht]
\caption{Common optimization settings.}
\label{tab:common-setup}
\centering
\small
\begin{tabular}{ll}
\toprule
Setting & Value \\
\midrule
Learning rate / weight decay & $10^{-6}$ / 0.01 \\
LR warmup ratio & 0 (Search main 3B: 0.1) \\
PPO epochs / loss reduction & 1 / token mean \\
Rollouts per prompt & 8 \\
Train / validation temperature & 1.0 / 0.4 (sampling) \\
Numeric type & bfloat16 \\
Gradient checkpointing & enabled \\
Actor / reference parameter offload & disabled / enabled \\
Teacher baseline / $\eta$ mode & step / fixed \\
Teacher temperature $T$ / gate $\tau$ & 1.0 / 2.0 \\
Score-normalization $\epsilon$ & $10^{-8}$ \\
GateNorm default & disabled \\
Score normalization & global or per-sequence \\
GiGPO $\gamma$ / step weight & 0.95 / 1.0 \\
GiGPO mode & mean normalization \\
Anchor similarity / threshold & disabled / 0.95 \\
\bottomrule
\end{tabular}
\end{table}

\begin{table}[!ht]
\caption{Primary optimization and environment settings.}
\label{tab:setup}
\centering
\scriptsize
\setlength{\tabcolsep}{3pt}
\begin{tabular}{lccc}
\toprule
Setting & ALFWorld & WebShop & Search \\
\midrule
Train batch & 16 & 16 & 128 \\
Eval batch & 128 & 128 & 1024 \\
Max turns & 50 & 15 & 4 \\
Prompt length & 2048 & 4096 & 4096 \\
Response length & 512 & 512 & 512 \\
Truncation & error & error & left \\
PPO mini-batch & 256 & 64 & 256 \\
Micro-batch / GPU & 16 & 8 & 16 \\
Invalid-action penalty & 0.1 & 0.1 & 0.01 \\
$\eta$ & 0.1 & 0.02 & 0.05--0.1 \\
KL coefficient & 0.01 & 0.01 & 0.001 \\
\bottomrule
\end{tabular}
\end{table}

All ADRS models are trained on multiple NVIDIA H100 GPUs. We follow the SDAR experimental configuration~\cite{lu2026self} for the benchmark interface, rollout schedule, and evaluation protocol, while replacing its optimization objective with the proposed reward-shaping framework.

\emph{ADRS-global} uses step centering and batch-global scaling; \emph{ADRS-per-seq} changes only score scaling to per-sequence normalization in the matched fix3/fix4 comparisons. \emph{GateNorm} applies Eq.~\ref{eq:tva}; \emph{ADRS (no TVA)} uses the corresponding environment configuration with $m=1$. The primary ALFWorld row uses GRPO, L2+GateNorm, $\eta=0.1$, entropy coefficient 0.01, asymmetric clip bounds 0.2/0.28, and gradient clipping at 1.0. The primary WebShop row uses GRPO, L2 per-sequence normalization, and $\eta=0.02$. The strongest 3B Search row at step 150 is the global-normalized $\eta=0.05$ run. The ALFWorld no-TVA/TVA rows in the component sequence instead share GiGPO and common optimization settings except the gate.

\subsection{TVA Levels and Non-Return Fallback}
\label{app:tva-levels}

The implementation chooses one available level for an entire run; it does not fall back separately for each undersized group.
\begin{itemize}
    \item \textbf{Step-level TVA (L3):} units are repeated GiGPO anchor states, and $R_u$ is the pre-ADRS environment return-to-go.
    \item \textbf{Completion-level TVA (L2):} units are trajectories sampled for the same prompt, and $R_u$ is the pre-ADRS completion return.
    \item \textbf{Token fallback (L1):} when reference log-probabilities exist but L2/L3 tensors do not, the implementation uses
    \begin{equation}
    m_t^{\mathrm{L1}}
    =\sg\!\left[\sigma\!\left(\tau(\ell^T_t-\ell_t^{\mathrm{ref}})\right)\right],
    \label{eq:l1-fallback}
    \end{equation}
    where $\ell_t^{\mathrm{ref}}$ is the detached frozen-reference score without skill text. This fallback contains no return and is not a TVA estimator.
\end{itemize}
At L2/L3, a group with fewer than two units is assigned $d_g=0$ and hence $m=0.5$; it does not dynamically fall back to another level. Turning TVA off is different: it sets $m=1$.

\subsection{Canonical Training Procedure}
\label{app:algorithm}

\paragraph{Algorithm 1: ADRS training iteration.}
\textbf{Input:} behavior policy $\pi_{\theta_b}$, task batch, skill provider $\rho$, teacher scale $\eta$, selected gate level, and backbone $\mathcal B$. \textbf{Output:} updated actor parameters $\theta$.
\begin{enumerate}
    \item Sample $K$ ordinary-context trajectories per task from $\pi_{\theta_b}$ and record valid-token masks, environment outcomes, and the backend's base token rewards.
    \item With $\theta_b$ frozen, compute ordinary scores $\ell^b$ and skill-conditioned scores $\ell^T$ on the same sampled tokens. Treat both score tensors and all subsequent teacher-branch quantities as detached.
    \item Compute the step-centered and normalized teacher score $\widehat q$ with Eq.~\ref{eq:teacher-score}.
    \item If TVA is enabled, compute the selected L2/L3 return contrast and detached gate with Eqs.~\ref{eq:tva-confidence}--\ref{eq:tva}; if L1 is selected, use Eq.~\ref{eq:l1-fallback}; otherwise set $m=1$.
    \item Form and save the teacher token reward $r^T=\eta m\widehat q$, and form the shaped token reward $\widetilde r=r^{\mathrm{base}}+r^T$ as in Eq.~\ref{eq:star-reward}.
    \item \textbf{Trajectory path:} use the backbone's native credit construction on $\widetilde r$ to obtain $A^{\mathrm{traj}}$. For the step-shared no-TVA/L2/L3 cases, the teacher reward has zero within-step sum, so this trajectory score equals the base-reward trajectory score. For GiGPO, retain its native step-level credit branch unchanged.
    \item \textbf{Token path:} whiten the saved $r^T$ over valid tokens within each prompt group $g$ to obtain the detached token modulation
    $Z^T_t=(r^T_t-\mu_g^T)/\sqrt{(\sigma_g^T)^2+\epsilon}$.
    \item Merge the two paths once at token level,
    $A^{\mathrm{ADRS}}_{i,k,t}=A^{\mathrm{traj}}_{i,k}+\eta Z^T_{i,k,t}$,
    and detach the combined coefficient. For GiGPO, add $\eta Z^T$ to the episode-level advantage before combining it with the unchanged native step-level credit.
    \item Optimize Eq.~\ref{eq:star-objective} with the backend's configured clipping, KL, entropy, and gradient-control terms unchanged. Within the ADRS actor term, only the current actor log-probability receives gradient; native entropy, KL, and other regularizers retain their configured gradients.
    \item Commit the updated actor as the behavior policy for the next iteration.
\end{enumerate}

\section{Proofs and Scope of the Identities}
\label{app:gradient}

\paragraph{Proof of Proposition~\ref{prop:centered-potential}.}
By the definition of $b_s$,
$\sum_{j\in\mathcal V_s}q_{s,j}=\sum_j\ell^T_{s,j}-|\mathcal V_s|b_s=0$.
Division by a positive scalar shared within the step preserves the zero sum, as does multiplication by the step-shared $\eta m_s$; hence
$\sum_{j\in\mathcal V_s}r^T_{s,j}=0$.
For any ordering $(j_1,\ldots,j_{n_s})$, define $\Phi_{s,1}=0$ and
$\Phi_{s,r+1}=\Phi_{s,r}+r^T_{s,j_r}$.
Then $r^T_{s,j_r}=\Phi_{s,r+1}-\Phi_{s,r}$ directly, and the endpoint is
$\Phi_{s,n_s+1}=\sum_{r=1}^{n_s} r^T_{s,j_r}=0$.
The claim covers the canonical no-TVA, L2, and L3 cases when the gate is constant over the same valid-token set used for centering. It preserves only the unweighted shaping mass on that sampled step; it is neither a discounted-return guarantee nor a fixed, path-independent Markov potential.

\paragraph{Proof of Proposition~\ref{prop:tva-covariance}.}
Write $\bar R_g=\mathbb E_g[R]$. With $\epsilon=0$,
\begin{align*}
d_g
&=\frac{\mathbb E_g[\alpha R]}{\bar\alpha_g}
-\frac{\mathbb E_g[(1-\alpha)R]}{1-\bar\alpha_g}\\
&=\frac{\mathbb E_g[\alpha R]-\bar\alpha_g\bar R_g}
{\bar\alpha_g(1-\bar\alpha_g)}
=\frac{\operatorname{Cov}_g(\alpha,R)}
{\bar\alpha_g(1-\bar\alpha_g)}.
\end{align*}
The denominator is positive under the stated condition. GateNorm either divides by another positive scalar or leaves $d_g$ unchanged, and the sigmoid with $\tau>0$ is strictly increasing, proving the sign statement. With finite $\epsilon$, the implemented contrast is a stabilized approximation to the covariance identity. Its sign and magnitude approach Eq.~\ref{eq:tva-covariance} when both group-weight sums dominate $\epsilon$; no exact sign equivalence is claimed in the degenerate regime.

\paragraph{Proof of Proposition~\ref{prop:unified-gradient}.}
Under the exact old-logprob condition $\ell^b_t=\log\pi_{\theta_b}(y_t\mid c_t)$ from Proposition~\ref{prop:unified-gradient}, at $\theta_b$ we have $\omega_t=1$ and
$\nabla_\theta\omega_t=\omega_t\nabla_\theta\log\pi_\theta
=\nabla_\theta\log\pi_\theta$.
The valid-token reduction, samples, $A_t^0$, $\Delta A_t$, and $a_t$ are fixed, so direct differentiation gives Eqs.~\ref{eq:star-score-update} and~\ref{eq:aux-score-update}. Setting $a_t=\Delta A_t$ makes their score-weighted gradients equal at that evaluation point.

The gradient equality is local. Away from the behavior-policy point, the ADRS coefficient is multiplied by $\omega_t$ whereas the auxiliary log-likelihood coefficient is not, and PPO clipping can separate them further. The result excludes full-vocabulary KL/JSD, representation matching, non-detached gates, post-combination advantage re-normalization, and an entire multi-epoch PPO update. It does not commute $f$ through $\operatorname{Adv}_{\mathcal B}$: ADRS's coefficient is explicitly $\Delta A=\operatorname{Adv}_{\mathcal B}(r^{\mathrm{base}}+f)-\operatorname{Adv}_{\mathcal B}(r^{\mathrm{base}})$.

\section{Supplementary Results}
\label{app:supplementary-results}

\subsection{Teacher Scale and Score Normalization}

Table~\ref{tab:eta} reports the Search development-set sweep over the teacher scale $\eta$. The best standard-checkpoint and late-training settings differ: $\eta=0.05$ is strongest at step 150, whereas $\eta=0.1$ reaches the highest observed late checkpoint. The $\eta=0.005$ run peaks early and falls to 9.6 at step 150, showing that an excessively weak teacher coefficient can produce unstable checkpoint behavior rather than a smooth reduction toward the outcome-only baseline.

\begin{table}[!ht]
\caption{Search development-set sensitivity to $\eta$ for ADRS-global+GateNorm.}
\label{tab:eta}
\centering
\small
\begin{tabular}{rrrrl}
\toprule
$\eta$ & Step 150 & Peak & Peak step & Behavior \\
\midrule
0.10 & 36.8 & \textbf{42.1} & 285 & Strong late rise \\
0.05 & \textbf{36.9} & 38.1 & 135 & Best at standard point \\
0.02 & 35.2 & 40.3 & 270 & Slower late rise \\
0.005 & 9.6 & 38.3 & 90 & Early peak, collapse \\
\bottomrule
\end{tabular}
\end{table}

Table~\ref{tab:normalization} compares global and per-sequence score normalization in matched fix3/fix4 pairs. Global normalization is stronger on ALFWorld and full Search, while per-sequence normalization is stronger on WebShop. This environment dependence explains why the main table uses benchmark-specific ADRS configurations rather than one universally selected normalization rule.

\begin{table}[!ht]
\caption{Best observed percentages for matched global (fix3) versus per-sequence (fix4) normalization runs. Each pair changes only \texttt{normalize\_mode}.}
\label{tab:normalization}
\centering
\small
\begin{tabular}{lrrl}
\toprule
Environment / setting & Global & Per-seq & Higher \\
\midrule
ALFWorld & 90.6 & 82.8 & Global \\
WebShop & 79.7 & 82.0 & Per-seq \\
Search, full & 40.5 & 38.4 & Global \\
\bottomrule
\end{tabular}
\end{table}

\subsection{Complete Long-Horizon and Ablation Matrices}

Table~\ref{tab:extended-full} expands the representative 300-step results in the main text to every named variant. Best, final, and last-three values distinguish peak attainment from late-training persistence. Table~\ref{tab:ablation-full} similarly separates fixed-step comparisons from peak and trajectory summaries, clarifying the contributions of ADRS, TVA, and the underlying backbone.

\begin{table*}[!t]
\caption{Complete 300-step result matrix. ``Best'' is the highest observed checkpoint of the named run; Search uses the full 51K evaluation at step 300. Rows characterize named variants rather than one universal configuration.}
\label{tab:extended-full}
\centering
\footnotesize
\begin{tabular}{llrrrl}
\toprule
Environment & Configuration & Best & Final & Last-3 & Interpretation \\
\midrule
\multirow{6}{*}{ALFWorld}
 & GRPO & 92.2 & 86.7 & 86.2 & Local outcome-only baseline \\
 & GiGPO & 96.9 & 91.4 & 93.0 & Strong step-level baseline \\
 & ADRS (no TVA), GiGPO & 96.1 & 92.2 & 91.9 & Fixed teacher scale \\
 & ADRS+TVA, GiGPO & 96.9 & 94.5 & 94.3 & Stronger reported late checkpoints \\
 & ADRS-global+GateNorm, GRPO & 96.1 & 89.8 & 90.6 & Primary step-150 system \\
 & ADRS-global+entropy/clip, GRPO & \textbf{97.7} & \textbf{96.1} & \textbf{96.1} & Strongest late-stage run \\
\midrule
\multirow{5}{*}{WebShop}
 & GRPO & 78.9 & 78.9 & 75.3 & Local outcome-only baseline \\
 & ADRS (no TVA) & 82.8 & 82.8 & 80.2 & Fixed teacher scale \\
 & ADRS-per-seq (primary) & 82.0 & 78.9 & 77.3 & Step-150 primary run \\
 & ADRS-per-seq (late) & \textbf{84.4} & \textbf{84.4} & 77.9 & Highest final checkpoint \\
 & ADRS-per-seq+GateNorm & \textbf{84.4} & 82.0 & 78.9 & Peak at step 275 \\
\bottomrule
\end{tabular}
\end{table*}

\begin{table*}[!t]
\caption{Complete ablation and backbone matrix. Last-3 is the mean of the final three reported checkpoints.}
\label{tab:ablation-full}
\centering
\footnotesize
\begin{tabular}{llllrrr}
\toprule
Environment & Evaluation & Factor & Reference & Ref. & Variant & Delta \\
\midrule
\multicolumn{7}{l}{\textit{Fixed-step local comparisons}} \\
ALFWorld & Step 150 & ADRS on GiGPO & GiGPO & 82.8 & ADRS+TVA 89.8 & +7.0 \\
ALFWorld & Step 150 & Centered ADRS & GiGPO & 82.8 & ADRS (no TVA) 87.5 & +4.7 \\
ALFWorld & Step 150 & TVA & ADRS (no TVA) & 87.5 & ADRS+TVA 89.8 & +2.3 \\
ALFWorld & Step 150 & Primary system & Local GRPO & 84.4 & Primary ADRS 94.5 & +10.1 \\
WebShop & Step 150 & ADRS system & GRPO & 63.3 & ADRS-per-seq 76.6 & +13.3 \\
Search & Step 150 & ADRS & GRPO & 38.4 & ADRS-global 38.3 & -0.1 \\
\midrule
\multicolumn{7}{l}{\textit{Descriptive peak and trajectory summaries}} \\
ALFWorld & Peak & Backbone & GRPO & 92.2 & GiGPO 96.9 & +4.7 \\
ALFWorld & Peak & ADRS on GRPO & GRPO & 92.2 & ADRS-entropy/clip 97.7 & +5.5 \\
ALFWorld & Peak & TVA & ADRS (no TVA) & 96.1 & ADRS+TVA 96.9 & +0.8 \\
ALFWorld & Last-3 & TVA & ADRS (no TVA) & 91.9 & ADRS+TVA 94.3 & +2.4 \\
WebShop & Peak & ADRS & GRPO & 78.9 & ADRS (no TVA) 82.8 & +3.9 \\
WebShop & Peak & GateNorm & ADRS-per-seq & 82.0 & +GateNorm 84.4 & +2.4 \\
Search & Step 300 & ADRS-global & GRPO & 40.2 & ADRS-global 40.5 & +0.3 \\
\bottomrule
\end{tabular}
\end{table*}

\paragraph{Optimization dynamics.}
The selected ALFWorld ADRS curve crosses 70\% at approximately step 70, while the published SDAR curve crosses that threshold near step 130, corresponding to a 1.9$\times$ faster rise in terms of optimizer steps. Entropy trajectories also distinguish stable and aggressive variants: ADRS-global remains near 0.6 and ADRS+TVA ends near 0.34, whereas the entropy/clip variant rises to 8.63 and becomes unstable.

\subsection{Action--Object Token Diagnostic}

Table~\ref{tab:token-diagnostic} provides the exact values underlying Figure~\ref{fig:kt-kp-diagnostic}. The action-to-other magnitude ratio rises from 1.6 at initialization to 95.3 at step 150 and remains high at step 250. After this stage, task performance approaches saturation and the policy changes more slowly, which explains the shift in the object term and the smaller ratio at the final checkpoint.

\begin{table*}[!t]
\caption{Theory-oriented token diagnostic on four fixed ALFWorld trajectories. Values are signed log-probability differences, not KL terms.}
\label{tab:token-diagnostic}
\centering
\small
\begin{tabular}{rrrrr}
\toprule
Step & Action & Object & Other & $|\mathrm{Action}|/|\mathrm{Other}|$ \\
\midrule
0 & -0.2509 & 0.3089 & 0.1552 & 1.6 \\
50 & -0.5155 & 0.4150 & 0.1037 & 5.0 \\
100 & -0.5563 & 0.2354 & 0.0539 & 10.3 \\
150 & -0.8582 & 0.4591 & 0.0085 & 95.3 \\
200 & -0.9665 & 0.3516 & 0.0373 & 26.1 \\
250 & -0.9089 & 0.4475 & 0.0229 & 39.5 \\
300 & -1.3105 & -0.3664 & 0.5108 & 2.6 \\
\bottomrule
\end{tabular}
\end{table*}

\subsection{Training Dynamics across Models and Domains}

\begin{figure*}[p]
    \centering
    \includegraphics[width=\textwidth]{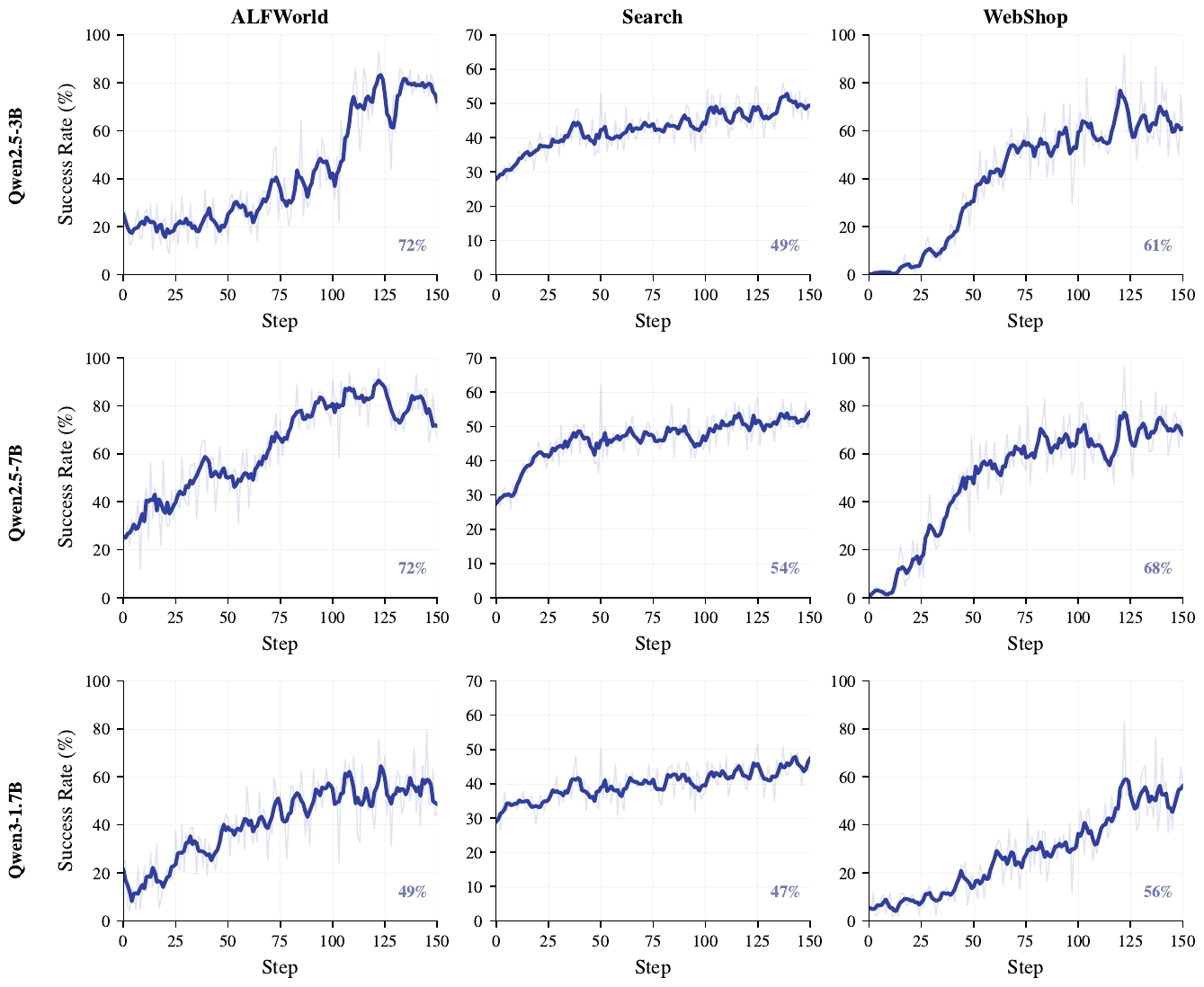}
    \Description{Nine panels of success-rate trajectories through step 150. Columns are ALFWorld, Search, and WebShop; rows are Qwen2.5-3B, Qwen2.5-7B, and Qwen3-1.7B. Light traces show checkpoint values and dark traces show smoothed trends.}
    \caption{ADRS success-rate dynamics across three environments and three model configurations. Light lines are checkpoint traces and dark lines are smoothed trends.}
    \label{fig:success-dynamics}
\end{figure*}

\begin{figure*}[p]
    \centering
    \includegraphics[width=\textwidth]{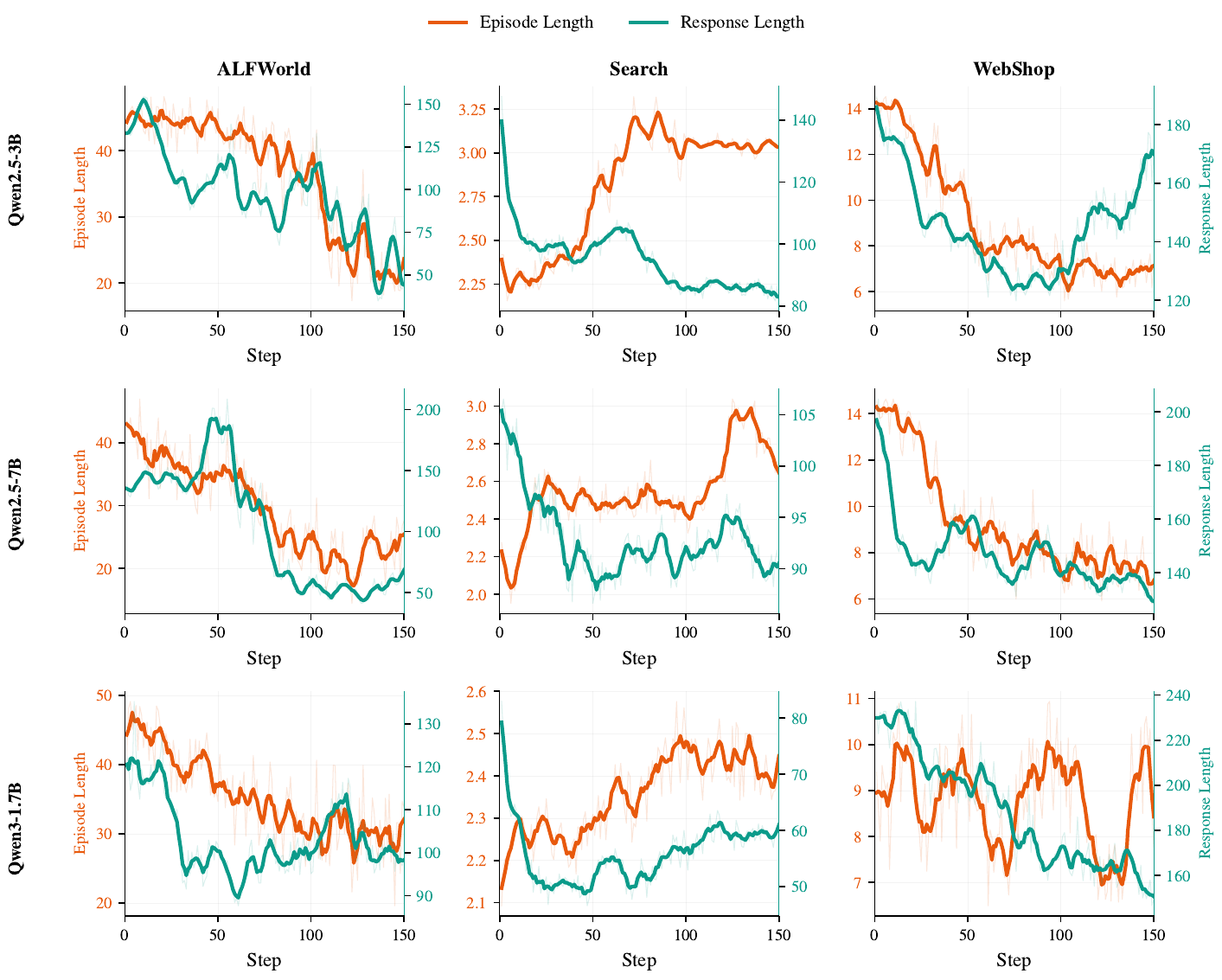}
    \Description{Two 3-by-3 grids of training trajectories through step 150. The upper grid shows mean episode length and the lower grid shows mean response length. Columns correspond to ALFWorld, Search, and WebShop; rows correspond to Qwen2.5-3B, Qwen2.5-7B, and Qwen3-1.7B.}
    \caption{Episode- and response-length dynamics across three environments and three model configurations. Top: mean episode length. Bottom: mean response length. Light lines show checkpoint traces, and dark lines show smoothed trends.}
    \label{fig:episode-length-dynamics}
\end{figure*}

Figures~\ref{fig:success-dynamics}--\ref{fig:episode-length-dynamics} show the full 3$\times$3 grid across Qwen2.5-3B, Qwen2.5-7B, and Qwen3-1.7B on ALFWorld, Search, and WebShop. Success generally rises over training, but the interaction statistics differ by environment: ALFWorld and WebShop often shorten episodes as success improves, whereas Search remains constrained to a small number of tool turns. Together, these trajectories show consistent optimization progress across model families and agentic domains.

\end{document}